\PassOptionsToPackage{table}{xcolor}
\documentclass{article} 
\usepackage[OT1]{fontenc} 
\usepackage{arxiv,times}

\usepackage{amsmath,amsfonts,bm}

\def\eqref#1{equation~\ref{#1}}

\def\1{\bm{1}}

\DeclareMathAlphabet{\mathsfit}{\encodingdefault}{\sfdefault}{m}{sl}
\SetMathAlphabet{\mathsfit}{bold}{\encodingdefault}{\sfdefault}{bx}{n}

\usepackage{multirow}
\usepackage{hyperref}
\usepackage{enumitem}
\usepackage{url}
\usepackage{graphicx}
\usepackage{amsmath}
\usepackage{amssymb}
\usepackage{booktabs}
\usepackage{tabularx} 
\usepackage{float} 
\usepackage{wrapfig}
\usepackage{caption}
\usepackage{placeins}

\title{LAMP: 3D Source-Aware Lighting Estimation at Multiple Points}

\author{Yanqiao Wang, Jin Sun\\
School of Computing\\
University of Georgia\\
Athens, GA 30602, USA \\
\texttt{\{yqiao.wang,jinsun\}@uga.edu} \\
}

\usepackage{xcolor}

\definecolor{oursgray}{gray}{0.90}

\newcommand{\oursrowprobe}[6]{%
& \cellcolor{oursgray}#1
& \cellcolor{oursgray}#2
& \cellcolor{oursgray}#3
& \cellcolor{oursgray}#4
& \cellcolor{oursgray}#5
& \cellcolor{oursgray}#6\\
}

\hypersetup{hidelinks}


\begin{document}

\maketitle

\begin{abstract}
Existing illumination estimation methods primarily recover lighting distributions for rendering, but typically leave the contributions of individual physical light sources at specific scene locations implicit. 
We address this gap by leveraging the scene understanding capabilities of vision-language models (VLMs) to connect local illumination with its physical sources.
We introduce LAMP, a framework for point-conditioned, source-aware illumination reasoning from a single RGB image with marked query points. LAMP constructs a shared inventory of active light sources, ranks contributing sources for each query, and predicts their 3D query-to-source directions.
We supervise these relationships using procedurally controlled indoor scenes, reusing independently rendered source responses to derive contribution targets across lighting configurations and obtaining exact directions from source and query positions.
On the Infinigen test set, fine-tuning Qwen3-VL-8B increases dominant-source 3D PCK@0.5\,m from 0\% to 70.6\%, with 94.5\% of directions within \(12.5^\circ\) of ground truth among correctly localized sources.
Combined with a lightweight network predicting dense RGB irradiance, LAMP achieves probe-rendering quality comparable to prior lighting-estimation methods while additionally localizing individual physical sources in 3D and explicitly attributing local illumination to them.
\end{abstract}

\section{Introduction}
\label{sec:intro}
In indoor scenes, illumination plays a central role in shaping appearance, and estimating it from images supports applications such as virtual object insertion and scene relighting. 
Single-image estimation is challenging because image intensities reflect the combined effects of illumination and surface reflectance, making their contributions difficult to disentangle~\citep{kajiya1986rendering}. 
Illumination also varies across locations: the same source can dominate one location while contributing little to another. Understanding these variations requires reasoning about the relationship between local lighting and the sources that produce it.

Existing methods estimate spatially varying illumination as local maps or basis coefficients~\citep{song2019neural,Garon_2019_CVPR,li2020inverse,bolduc2026lighting}, or as three-dimensional radiance representations~\citep{srinivasan2020lighthouse,wang2021learning,wang2024lightoctree,bolduc2025gaslight}. 
These representations support rendering but typically leave the relationship between local illumination and individual physical sources implicit. 
Explicit source identities and 3D positions instead provide a shared physical reference across locations.
This motivates recovering not only local lighting estimates, but also their connections to active physical sources in 3D. 
The recognizable semantic identities of lighting fixtures make vision-language models (VLMs) a natural candidate for establishing these connections. 
However, determining whether a recognized light fixture is emitting light and how it contributes to illumination at a queried point requires reasoning beyond object recognition, including physical and spatial reasoning that remains challenging for VLMs~\citep{chow2025physbench,kamath2023s,wang2024picture}.

\begin{figure*}[t]
    \centering
    
    \begin{minipage}[c]{0.56\textwidth}
        \centering
        \includegraphics[width=\linewidth]{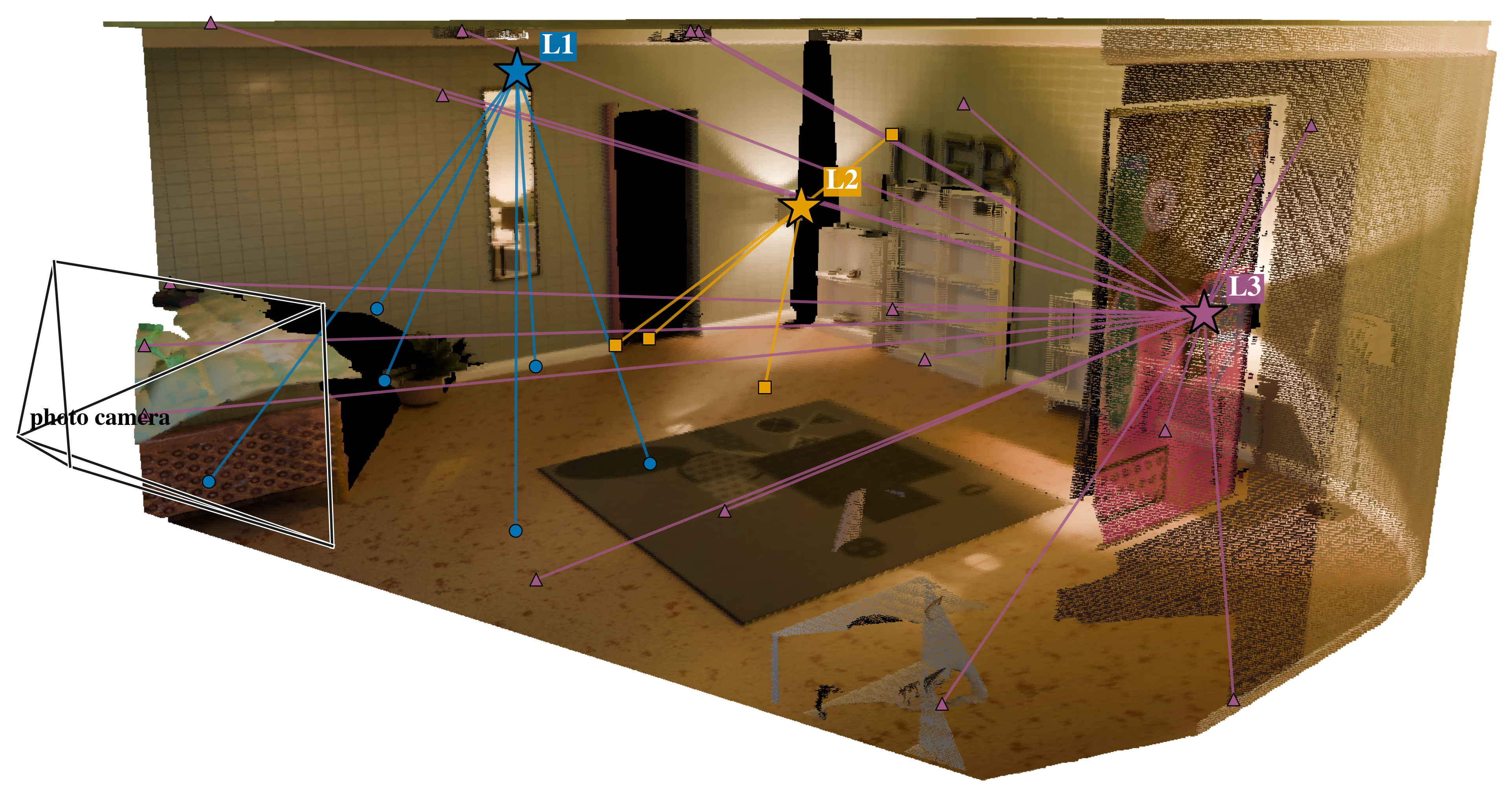}
    \end{minipage}
    \hfill
    \begin{minipage}[c]{0.34\textwidth}
        \centering


        \includegraphics[width=\linewidth]{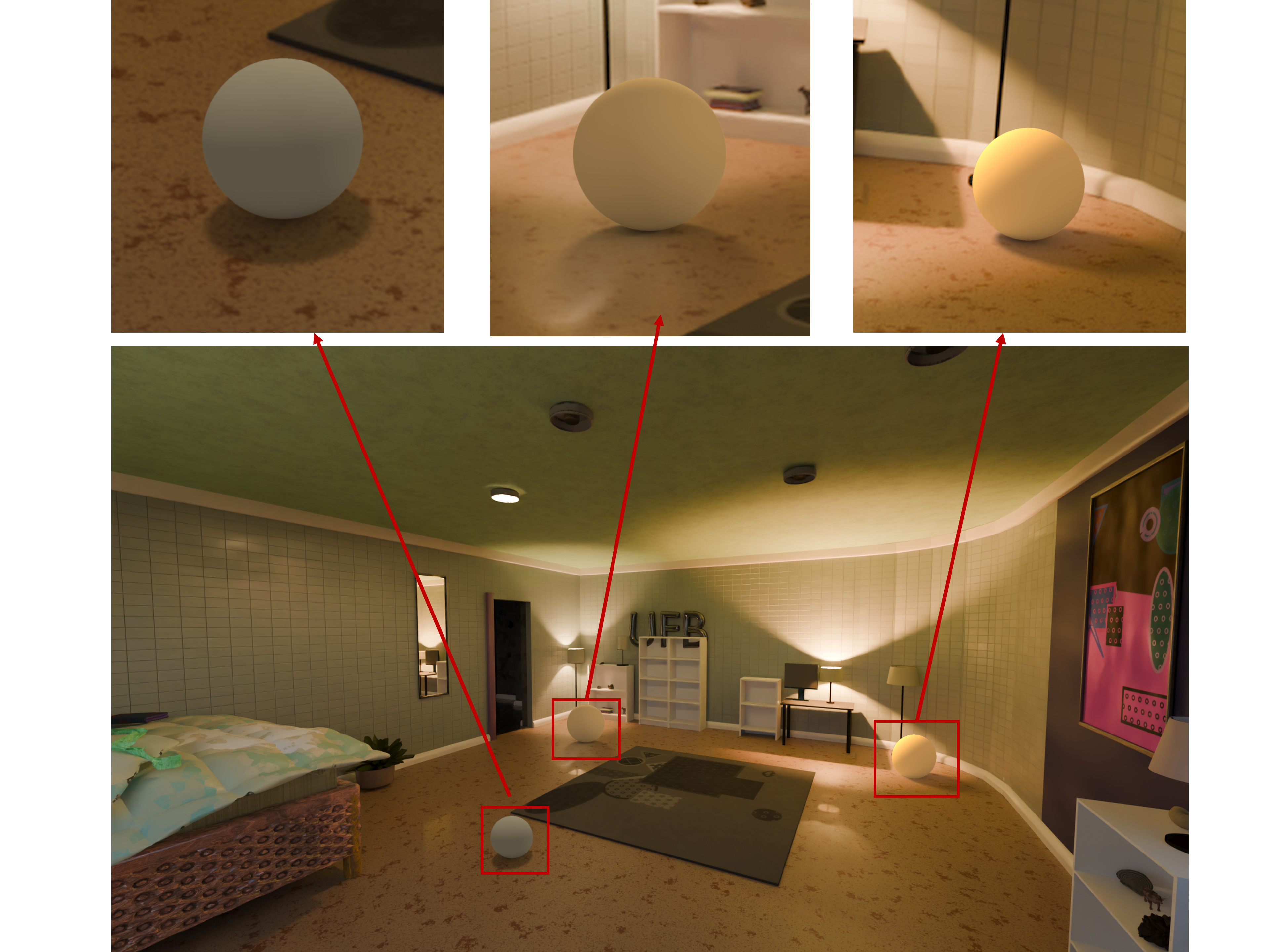}
    \end{minipage}

    \caption{\small{\textbf{Visualization of source-referenced illumination.} Left: predicted 3D source positions and dominant-source attribution, with colored lines connecting each query to its predicted dominant source. Right: probes rendered using the predicted source information.}}
    \label{fig:overview}
\end{figure*}

We introduce LAMP, a novel framework for source-aware illumination modeling that combines a shared 3D source inventory, query-conditioned ranked multi-source attribution, and a dense RGB irradiance field. 
Given a single RGB image marked with query points, a fine-tuned Qwen3-VL~\citep{bai2025qwen3} predicts a shared inventory of light sources with their 3D positions and ranks their contribution for each query, with the dominant source ranked first.
A complementary lightweight ResNet estimates dense RGB irradiance, providing local lighting magnitude and color. Sampling this field at the query pairs aggregate irradiance with the identity and direction of its dominant contributor. 
Together, these predictions connect local photometric estimates to explicit 3D sources without requiring complete recovery of each source's emission and transported contribution. Figure~\ref{fig:overview} shows the predicted source inventory, query-conditioned relationships, and their use for rendering.

To train this framework, we construct a procedurally controlled dataset using Infinigen Indoors~\citep{raistrick2024infinigen} and LuxRemix~\citep{Liang_2026_CVPR}. 
Rendering individual lights while holding scene structure, materials, and viewpoint fixed isolates each source's contribution. 
Through the linearity of light transport, these solo-source responses can be emission-normalized and recombined to generate contribution targets across multiple emission configurations without repeating solo-source rendering. 
Together, this data and training design link a shared source inventory to local contributions and capture changes in source dominance across locations and lighting configurations.
On the verified source-aligned subset of the Infinigen test set, our model achieves 96.6\% 3D query-to-source direction accuracy within \(22.5^\circ\), compared with at most 32.2\% for the evaluated lighting-estimation baselines.
While these methods estimate lighting without explicitly identifying its physical sources, our framework additionally identifies and localizes contributing sources in 3D while achieving probe-rendering quality comparable to existing lighting-estimation methods.

Our contributions are as follows:
\begin{itemize}[
    leftmargin=1.3em,
    topsep=0pt,
    itemsep=0pt,
    parsep=0pt,
    partopsep=0pt
]
\item We introduce a source-aware illumination representation that links local lighting to a shared inventory of physical sources in 3D, with query-conditioned source attribution, query-to-source directions, and dense RGB irradiance.

\item We construct a procedurally controlled dataset and an efficient supervision scheme that links individual sources to their local lighting contributions, reusing emission-normalized solo-source renders across lighting configurations.

\item We develop a multi-query prediction framework that combines VLM-based source reasoning with dense irradiance estimation, and uses training-only contribution-share supervision to capture spatially varying source influence beyond dominant-source labels.
\end{itemize}

\section{Related Work}

\paragraph{Illumination Estimation, Representation, and Relighting.}
Single-image illumination estimation has evolved from global HDR maps and parametric lights~\citep{gardner-sigasia-17,gardner2019deep} to local directional representations~\citep{Garon_2019_CVPR,song2019neural} and spatially coherent volumetric or hierarchical lighting~\citep{srinivasan2020lighthouse,wang2024lightoctree}. 
Recent methods introduce generative priors for HDR estimation~\citep{phongthawee2024diffusionlight,liang2026luxdit} and combine them with spatial representations to improve detail and consistency~\citep{bolduc2025gaslight,tong2025spatiotemporally,bolduc2026lighting}.
These advances support virtual object insertion and controllable relighting~\citep{choi2025scribblelight,Xing_2025_CVPR,magar2025lightlab}, including source-level editing through individual light decomposition~\citep{Liang_2026_CVPR}. 
Our complementary goal is to attribute local illumination at queried points to explicit scene sources and recover the corresponding point-to-source geometry.

\paragraph{Vision-Language Models for Lighting and Physical Reasoning.}
VLM-based illumination studies address robustness to lighting changes and color constancy~\citep{liu2025lighting,li2026white}, shared multimodal lighting representations~\citep{zhang2026unilight}, and dominant illumination estimation and description~\citep{hong2026single}. 
Other systems support lighting parameter editing~\citep{huang2024blenderalchemy,gu2025blendergym} or recover parametric sources from reconstructed 3D observations~\citep{chen2026engine}. These works primarily characterize global lighting or scene-level configurations, leaving source-specific effects at queried locations largely unexplored.
Broader benchmarks reveal persistent limitations in physical and spatial reasoning~\citep{chow2025physbench,balazadeh2025physics,kamath2023s,wang2024picture}, while targeted 3D supervision improves these capabilities~\citep{chen2024spatialvlm}. We study illumination as a grounded reasoning task combining source discovery, point-conditioned attribution, and continuous direction prediction.

\paragraph{Procedural Supervision for Grounded Physical Reasoning.}
Synthetic datasets and procedural generators provide rich indoor annotations and controllable scene geometry that are difficult to obtain from real images~\citep{roberts2021hypersim,li2021openrooms,greff2022kubric,raistrick2023infinite,raistrick2024infinigen}.
Such control supports spatial lighting supervision~\citep{tong2025spatiotemporally,bolduc2026lighting} and individual light decomposition~\citep{Liang_2026_CVPR}. We use source-specific rendered contributions and known geometry to supervise correspondences between light sources, their local effects, and point-to-source directions.
\section{Method}
\label{sec:method}
\begin{figure}[t]
    \centering
    \includegraphics[width=0.8\linewidth]{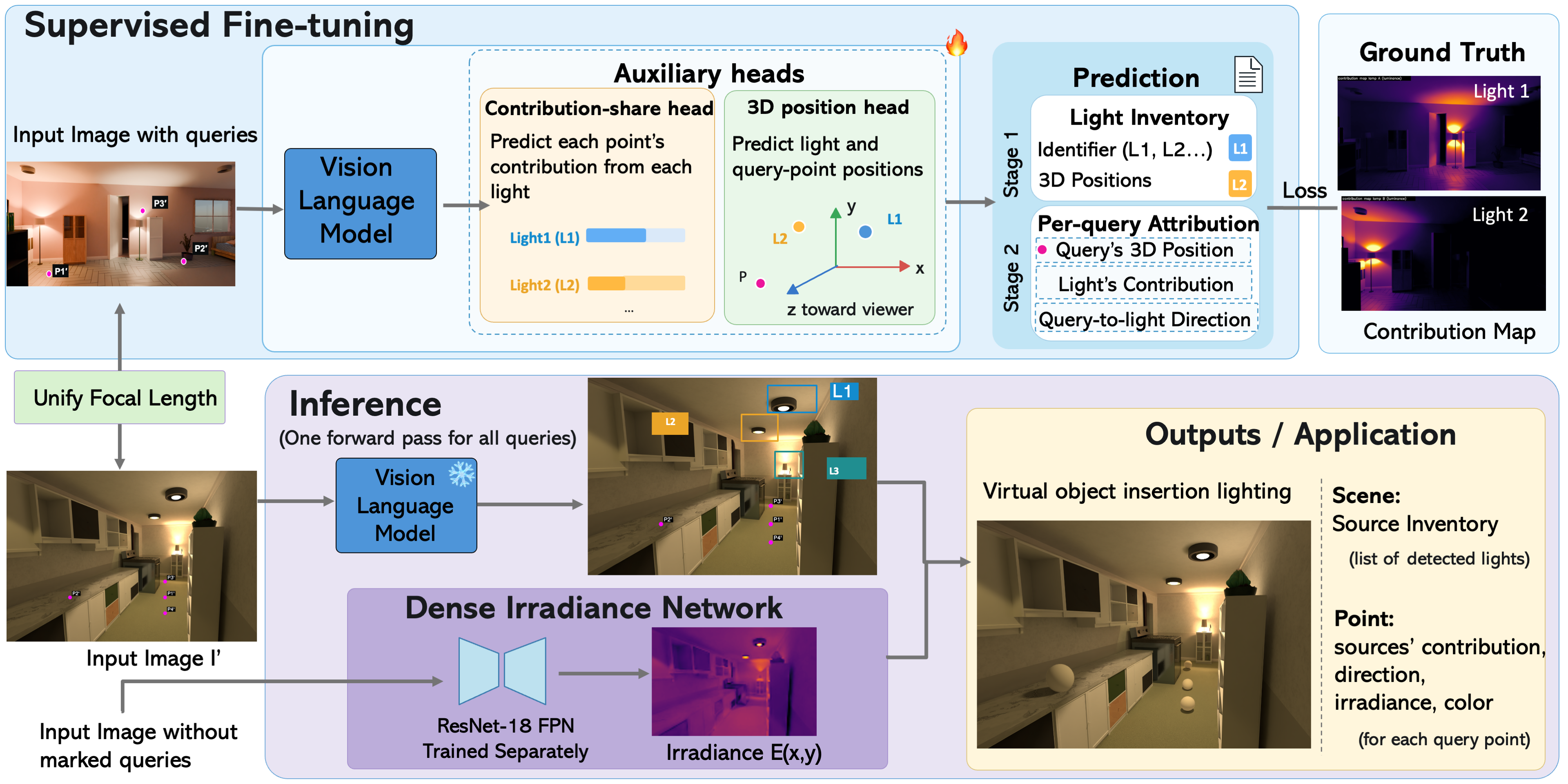}
    \caption{\small
    \textbf{Pipeline Overview.}
    Given the focal-length-normalized image with query markers and a text prompt, the VLM produces the active source inventory (Stage 1) and the query-conditioned records (Stage 2).
    During training, two auxiliary objectives, contribution-share attribution and position regression, act on the identifier tokens and are discarded afterwards.
    A dense network predicts RGB irradiance from the unmarked image.
    At inference, the ranked sources and the irradiance map are composed to render a probe at any query.
    }
    \label{fig:pipeline}
\end{figure}

Our framework combines structured source reasoning with dense RGB irradiance estimation (Figure~\ref{fig:pipeline}). 
Given an RGB image rescaled to a canonical focal length and then annotated with query markers, together with a text prompt, a fine-tuned Qwen3-VL generates a two-stage response. 
Stage 1 produces a scene-level inventory of active light sources with their 3D positions.
Stage 2 predicts each query's 3D position, the number of relevant sources, and a contribution-ranked list of these sources with their query-to-source directions, whose first entry is the predicted dominant source as defined in~\eqref{eq:share_and_dominance}.
A lightweight ResNet estimates dense RGB irradiance in one forward pass, providing aggregate lighting magnitude and color.

\subsection{Problem Formulation}
\label{sec:problem_formulation}

\paragraph{Input and structured output.}
Given an RGB image \(I\), query locations \(\mathcal{Q}=\{\mathbf{q}_i\}_{i=1}^{N_q}\) are annotated on the focal-length-normalized image with circular markers labeled \texttt{P1},\texttt{P2}, \ldots

The prompt specifies the number \(N_q\) of queried points without providing their numerical coordinates.
Under lighting configuration \(l\), \(\mathcal{S}^{l}=\{s_k\}_{k=1}^{K^{l}}\) denotes the active, individually controlled emitters, regardless of camera visibility, including partially visible, occluded, and off-screen sources.
Environment illumination, such as daylight through windows, is not a candidate source and is treated as a fixed background (Section~\ref{sec:per_light_transport}). 

The model first predicts an active source inventory \(\widehat{\mathcal{S}}^{l}\) whose entries contain an inventory-local identifier (\texttt{L1}, \texttt{L2}, \ldots), a pixel location (\texttt{null} if off-screen), and a 3D position.
Switched-off fixtures are excluded, and identifiers are local to each example rather than externally supplied (Appendix~\ref{app:problem_inventory_identifier}).

For each query \(\mathbf{q}_i\), the model additionally predicts its 3D position \(\widehat{\mathbf{P}}_i\in\mathbb{R}^{3}\), relevant-source count \(\widehat{n}_i\), and a ranked list of the sources that contribute to its illumination, written as

\begin{equation}
\widehat{\mathbf{y}}_i
=
\Bigl(
\widehat{\mathbf{P}}_i,
\widehat{n}_i,
\bigl\{(\widehat{s}_{i,r},\,
\widehat{\boldsymbol{\omega}}_{i,r},\,
\widehat{\boldsymbol{\omega}}_{i,r}^{\mathrm{2D}},\,
\widehat{w}_{i,r})\bigr\}_{r=1}^{\widehat{m}_i}
\Bigr),
\qquad
\widehat{m}_i=\min(\widehat{n}_i,N_D),
\quad N_D=4
\label{eq:structured_query_output}
\end{equation}

where each entry references an inventory source \(\widehat{s}_{i,r}\in\widehat{\mathcal{S}}^{l}\) through its inventory-local identifier, \(\widehat{\boldsymbol{\omega}}_{i,r}\in\mathbb{R}^{3}\) is the unit query-to-source direction in the camera frame, \(\widehat{\boldsymbol{\omega}}_{i,r}^{\mathrm{2D}}\) is its image-plane counterpart (the unit vector from the marker pixel to the projected source pixel), and 
 \(\widehat{w}_{i,r}\in[0,1]\) is the predicted contribution share of that source at the query. 
Entries are ranked by decreasing share, and the first entry is the predicted dominant source, defined in~\eqref{eq:share_and_dominance}. 
Ground-truth shares and the relevance threshold are defined in Section~\ref{sec:per_light_transport}.

\paragraph{Camera-frame targets.}
All 3D positions and directions use the input camera frame, with \(x\) right, \(y\) up, and \(z\) toward the viewer.
Renderer world coordinates are transformed using the inverse camera-to-world pose.
For source \(s_k\), \(\mathbf{X}_k\) denotes the emitter origin for point and spot lights or the center for area lights. The direction target is the unit vector from the query position to \(\mathbf{X}_k\).
The pose is used only to construct targets and is never provided to the model.

\paragraph{Focal-length normalization.}
To account for image-scale variation caused by focal length, we resample each image to a common pixel focal length \(f_{\mathrm{uni}}\) before drawing the query markers, with
\begin{equation}
I'=\operatorname{resize}(I,\gamma_{\mathrm{resize}}),
\qquad
\gamma_{\mathrm{resize}}=\frac{f_{\mathrm{uni}}}{f_x},
\qquad
\mathbf{q}'_i=\gamma_{\mathrm{resize}}\mathbf{q}_i.
\end{equation}
All image-space annotations, including query markers and source pixel locations, are transformed into the coordinate system of \(I'\), while metric 3D positions and directions remain unchanged.
This standardizes the relation between pixel offsets and geometry at a given depth. 
The same preprocessing is applied at inference using the input image's focal length.
The focal length is used only for this resampling and is not included in the model prompt.

\subsection{Emission-Normalized Per-Light Transport}
\label{sec:per_light_transport}

\begin{figure}[H]
    \centering
    \includegraphics[width=0.92\linewidth]{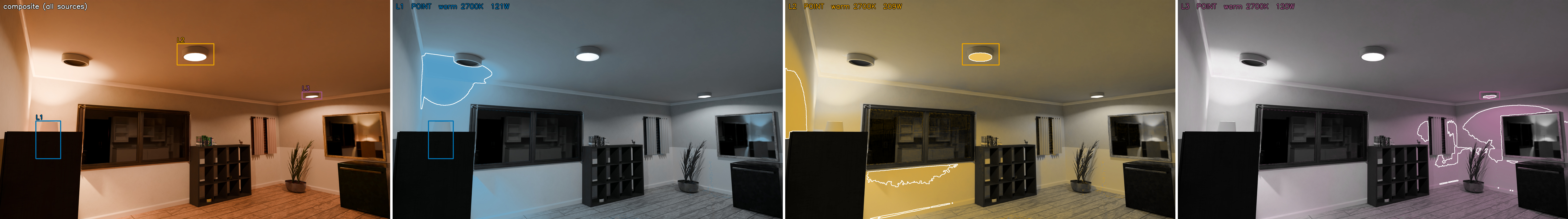}
    \caption{\small
    \textbf{Per-source contributions.} The composite render is shown with solo-source responses used to recover each active source's emission-normalized contribution \(\mathbf{C}_{s_k}^{l}\). Contributions are overlaid on the grayscale composite, with contours marking the top 3\% of values. Overlays are for visualization only; share labels use the continuous linear-RGB maps.}
    \label{fig:per_light_contributions}
\end{figure}

\paragraph{Per-light transport basis.}
Our targets require the contribution of every source at every pixel under many lighting configurations. 
Rendering each configuration once per isolated source would multiply the rendering cost by the number of configurations. 
We avoid this by exploiting the linearity and superposition of light transport with respect to emitted radiance~\citep{kajiya1986rendering,veach1997robust}.
With geometry, materials, and camera fixed, only the emission of each source varies between configurations, and the image radiance under configuration \(l\) decomposes as
\begin{equation}
\mathbf{L}^{l}(\mathbf{u})=\mathbf{L}_{\mathrm{off}}(\mathbf{u})+\sum_{k=1}^{K^{l}}\boldsymbol{\eta}_{s_k}^{l}\odot\mathbf{T}_{s_k}(\mathbf{u}),
\label{eq:linear_light_configuration}
\end{equation}
where \(\boldsymbol{\eta}_{s_k}^{l}\in\mathbb{R}^{3}_{\geq 0}\), non-zero for an active source, is the RGB emission of source \(s_k\), \(\mathbf{T}_{s_k}\) its unit-emission transport response (visibility, shadows, reflection, and inter-reflection included), and \(\mathbf{L}_{\mathrm{off}}\) the all-controlled-emitters-off baseline. 
We therefore render each source once, alone, under a reference emission \(\boldsymbol{\eta}_{s_k}^{\mathrm{ref}}\), giving \(\mathbf{L}_{s_k}^{\mathrm{ref}}\). 
Its contribution under any configuration \(l\) follows by channel-wise emission normalization as
\begin{equation}
\mathbf{C}_{s_k}^{l}(\mathbf{u})=\bigl(\boldsymbol{\eta}_{s_k}^{l}\oslash\boldsymbol{\eta}_{s_k}^{\mathrm{ref}}\bigr)\odot\bigl[\mathbf{L}_{s_k}^{\mathrm{ref}}(\mathbf{u})-\mathbf{L}_{\mathrm{off}}(\mathbf{u})\bigr].
\label{eq:emission_normalized_contribution}
\end{equation}
Figure~\ref{fig:per_light_contributions} visualizes the per-source contributions computed from the reference solo renders for a composite lighting configuration.
One transport basis per view therefore supervises any configuration obtained by scaling the reference emissions of individual sources, including switching them off, dimming them, or brightening them while everything else remains fixed. 
Each configuration is rendered directly for the input image. See Appendix~\ref{app:emission_transport_basis} for implementation details.

\paragraph{Contribution shares and ranked targets.}
Source attribution is defined by each source's contribution to the camera-observed radiance at the query point.
For query \(\mathbf{q}_i\) and the active sources \(\mathcal{S}^{l}\), the share of source \(s_k\) is the fraction of the total source luminance it contributes, and the dominant source is the source with the largest share, that is 
\begin{equation}
r_{ik}^{l}=\frac{\operatorname{lum}\bigl(\mathbf{C}_{s_k}^{l}(\mathbf{q}_i)\bigr)}{\sum_{j=1}^{K^{l}}\operatorname{lum}\bigl(\mathbf{C}_{s_j}^{l}(\mathbf{q}_i)\bigr)},
\qquad
k_i^{l,*}=\underset{1\leq k\leq K^{l}}{\arg\max}\;r_{ik}^{l},
\label{eq:share_and_dominance}
\end{equation}
with \(\operatorname{lum}\) the linear luminance.
The dominant source is defined by~\eqref{eq:share_and_dominance} throughout the paper, and the baseline \(\mathbf{L}_{\mathrm{off}}\) does not enter the normalization. 
The full share vector supervises the attribution head (Section~\ref{sec:learning_framework}). 
The relevant-source count \(n_i^{l}\) is the number of sources with \(r_{ik}^{l}\geq\tau_{\mathrm{share}}=0.05\). 
The ranked target lists these sources in decreasing order of share and is truncated to \(\min(n_i^{l},N_D)\) entries. 

\subsection{Structured Learning with Contribution-Share Supervision}
\label{sec:learning_framework}
We initialize structured training from Qwen3-VL-8B fine-tuned on single-marker direction prediction using the same rendered training views, then update all model parameters.
Training combines the main two-stage attribution examples with auxiliary text tasks for 3D positions, full contribution shares, and matched lighting configurations.
All examples receive token-level cross-entropy supervision over their task-specific JSON targets, while the two auxiliary heads below are applied only to two-stage attribution examples.

\paragraph{Structured autoregressive supervision.}
Each example pairs an image with 1-40 query markers, a prompt, and a task-specific JSON target.
For two-stage attribution examples, the target contains the source inventory followed by the query-conditioned records of~\eqref{eq:structured_query_output}, with counts and ranked entries.
The auxiliary examples isolate complementary supervision for 3D positions, untruncated contribution shares, or changes in query records across matched lighting configurations.
We apply token-level cross-entropy, denoted \(\mathcal{L}_{\mathrm{CE}}\), to the complete target sequence.

\paragraph{Contribution-share attribution.}
The ranked target includes only sources with share at least \(\tau_{\mathrm{share}}\), up to \(N_D\) entries.
To supervise contributions beyond this thresholded and truncated list, we add a training-only pointer head on two-stage attribution examples.
Under teacher forcing, the head reads the final-layer hidden state at the last subtoken of each query identifier and each source identifier in the inventory.
Two learned linear maps project the query and source states into a shared \(d\)-dimensional space.
A softmax over their scaled dot products yields a distribution \(\mathbf{p}_i^{l}\) over the \(K^{l}\) inventory sources for query \(\mathbf{q}_i\) under configuration \(l\).
We supervise this distribution with the full share vector \(\mathbf{r}_i^{l}\) from~\eqref{eq:share_and_dominance}, without thresholding or truncation, using \(\mathcal{L}_{\mathrm{attr}}=\mathbb{E}_{(i,l)}\!\left[D_{\mathrm{KL}}\!\left(\mathbf{r}_i^{l}\,\|\,\mathbf{p}_i^{l}\right)\right]\), where the expectation is over queries in attribution examples. Share-only examples carry their full share vectors as text and receive only \(\mathcal{L}_{\mathrm{CE}}\) supervision.

\paragraph{Auxiliary position supervision.}
On two-stage attribution examples, two MLP heads predict source and query positions in camera-frame 3D coordinates from their respective identifier states.
The loss \(\mathcal{L}_{\mathrm{reg}}\) is the mean absolute error against the renderer-derived positions over all source and query coordinates within each example, then averaged over attribution examples.

\paragraph{Overall objective.}
The complete training objective is \(\mathcal{L}=\mathcal{L}_{\mathrm{CE}}+0.5\,\mathcal{L}_{\mathrm{attr}}+0.1\,\mathcal{L}_{\mathrm{reg}}\).
All auxiliary heads are removed after training.

\paragraph{Inference and count-consistent decoding.}
At inference, the marked image and prompt are greedily decoded in one pass.
A count-consistent fallback can append missing ranked entries using unused inventory identifiers, but it is never triggered on either held-out split.
Implementation details are provided in the appendix~\ref{app:constrained_decoding}.

\subsection{Dense Irradiance Map}
\label{sec:rgb_irradiance}
The structured output describes which sources contribute to a query and from where, but not how much light arrives.
We therefore use a lightweight ResNet to predict a low-resolution map of log RGB irradiance from the unmarked image, sampled at each query.
Its supervision is per-pixel surface irradiance derived from renderer diffuse passes with albedo divided out, providing aggregate magnitude and color complementary to source attribution.
Probe calibration, color augmentation, architecture, and composition details are provided in Appendix~\ref{app:irradiance_details}.

\section{Experiments}
\label{sec:experiments}

\subsection{Experimental Setup}
\label{sec:experimental_setup}

\paragraph{Data.}
We train on Infinigen Indoors~\citep{raistrick2024infinigen} and LuxRemix~\citep{Liang_2026_CVPR}, totaling 498 scenes, 10{,}330 images, and 51{,}890 attribution queries.
Training, validation, and test splits are scene-disjoint.
Probe rendering additionally uses Laval Indoor HDR~\citep{gardner-sigasia-17} (Laval-RF) and Laval-SV~\citep{Garon_2019_CVPR}, while Hypersim~\citep{roberts2021hypersim} is used only qualitatively.

\paragraph{Models and baselines.}
We compare our fine-tuned Qwen3-VL-8B-Instruct (Section~\ref{sec:method}) with Qwen3-VL-8B (zero-shot) using the same structured prompt.
For source inventory evaluation, we fine-tune YOLOv8n~\citep{yolov8_ultralytics} on our training images and lift its detections to 3D using either ground-truth depth or Depth Anything~v2 (DAv2)~\citep{depth_anything_v2}.
Direction baselines include IllumiDiff~\citep{shen2025illumidiff}, DiffusionLight-Turbo~\citep{Chinchuthakun2025DiffusionLightTurbo}, and LiMo~\citep{bolduc2026lighting}, with directions extracted from their predicted lighting peaks.
For probe rendering, we compare with LiMo, DiffusionLight-Turbo, LuxDiT~\citep{liang2026luxdit}, GaSLight~\citep{bolduc2025gaslight}, and the method of~\citep{Garon_2019_CVPR} using the authors' released predictions on Laval-SV.

\paragraph{Metrics.}
\textbf{3D~PCK@\(\tau\)} requires the inventory position referenced at rank \(r\) to lie within \(\tau\) meters of the corresponding ground-truth ranked source and to have that source as its nearest neighbor among all ground-truth active sources.
We use \(\tau\in\{0.25,0.5,1.0\}\)\,m, where 0.25\,m is comparable to a fixture's extent and 0.5\,m is below typical inter-source spacing in our scenes.
Rank \(r\) evaluates the \(r\)-th ranked source on queries with at least \(r\) ground-truth relevant sources. Rank 1 is the dominant source.
Unparsed, missing, or unlocated entries count as incorrect.
\textbf{3D Direction@\(\theta\)} measures the fraction of predictions within \(\theta\in\{12.5^\circ,22.5^\circ,30^\circ\}\) of the corresponding ground-truth query-to-source direction, with missing or invalid directions counted as incorrect.
\textbf{Dir\(\mid\)PCK} restricts this evaluation to PCK@0.5\,m-correct entries, and we also report \textbf{median angular error} over valid predictions.
For \textbf{probe rendering}, we compare rendered and reference diffuse spheres using LPIPS, RMSE, si-RMSE, SSIM, and RGB angular error over sphere pixels.
Infinigen and Laval-SV evaluate queried locations directly, whereas LuxRemix and Laval-RF provide camera-position references, for which local estimates are transferred from the query nearest the image center.

\begin{table}[!t]
\centering
\caption{\textbf{Source localization and query-to-source direction.}
Rank 1 denotes the dominant source. Direction accuracy (\%) and median error are conditioned on PCK@0.5\,m; ``--'' denotes unavailable or undefined results. Median angular error here is computed on valid predictions.}
\label{tab:main_source_referenced}
\small
\setlength{\tabcolsep}{2.5pt}
\renewcommand{\arraystretch}{1.04}
\begin{tabularx}{\linewidth}{@{}l r *{6}{>{\centering\arraybackslash}X}c@{}}
\toprule
Model / Rank & \(N_q\)
& \multicolumn{3}{c}{3D PCK (\%) $\uparrow$}
& \multicolumn{4}{c}{Direction $\mid$ PCK@0.5\,m} \\
\cmidrule(lr){3-5}\cmidrule(lr){6-9}
& & 0.25\,m & 0.5\,m & 1.0\,m
& $12.5^\circ\!\uparrow$ & $22.5^\circ\!\uparrow$ & $30^\circ\!\uparrow$ & Med. ($^\circ$)$\downarrow$ \\
\midrule
\multicolumn{9}{@{}l}{\textbf{Infinigen}} \\
Qwen3-VL-8B (zero-shot) & 1{,}032 & 0.0 & 0.0 & 0.0 & -- & -- & -- & -- \\
\rowcolor{oursgray}
Ours, rank 1 & 1{,}032 & \textbf{49.2} & \textbf{70.6} & \textbf{82.2}
& \textbf{94.5} & \textbf{98.2} & \textbf{98.8} & \textbf{2.7} \\
Ours, rank 2 & 586 & 31.4 & 49.5 & 56.8 & 97.6 & 99.0 & 99.3 & 2.1 \\
Ours, rank 3 & 169 & 32.0 & 40.2 & 46.2 & 92.6 & 97.1 & 98.5 & 2.6 \\
\midrule
\multicolumn{9}{@{}l}{\textbf{LuxRemix}} \\
Qwen3-VL-8B (zero-shot) & 9{,}412 & 0.0 & 0.1 & 0.4 & -- & -- & -- & -- \\
\rowcolor{oursgray}
Ours, rank 1 & 9{,}412 & \textbf{10.1} & \textbf{28.5} & \textbf{46.0}
& \textbf{65.7} & \textbf{87.6} & \textbf{93.9} & \textbf{9.1} \\
Ours, rank 2 & 8{,}538 & 3.2 & 9.7 & 18.5 & 77.2 & 93.9 & 97.0 & 7.6 \\
Ours, rank 3 & 6{,}258 & 2.0 & 6.2 & 11.8 & 82.3 & 96.7 & 99.5 & 6.9 \\
Ours, rank 4 & 3{,}357 & 0.6 & 2.1 & 3.3 & 88.4 & 100.0 & 100.0 & 6.2 \\
\bottomrule
\end{tabularx}
\end{table}

\subsection{Experiment Results}
\label{sec:experiment_results}

\paragraph{Source localization and direction.}
Because our representation is explicitly source-referenced, we first evaluate whether the model can recover contributing sources and their 3D query-to-source directions, including secondary contributors.
Table~\ref{tab:main_source_referenced} shows strong dominant-source recovery after fine-tuning, while conditional direction remains accurate for successfully localized higher-rank sources. Full direction and scene-condition breakdowns are provided in Appendix~\ref{app:single_multi}.

\begin{figure}[!htbp]
\centering
\includegraphics[width=\linewidth]{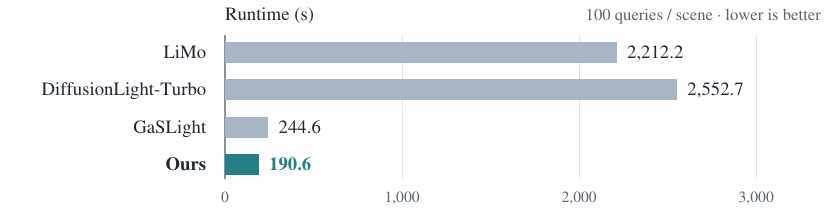}
\caption{\small{\textbf{Inference cost for 100 queries.}
Sharing predictions across queries substantially reduces runtime.}}
\label{fig:runtime}
\end{figure}

\begin{figure}[!t]
\centering
\begin{minipage}[c]{0.018\textwidth}\mbox{}\end{minipage}%
\hspace{0.012\textwidth}%
\begin{minipage}[c]{0.965\textwidth}\scriptsize\noindent
\begin{minipage}[c]{0.16666\linewidth}\centering Input\end{minipage}%
\begin{minipage}[c]{0.16666\linewidth}\centering LiMo\end{minipage}%
\begin{minipage}[c]{0.16666\linewidth}\centering DiffusionLight-\\Turbo\end{minipage}%
\begin{minipage}[c]{0.16666\linewidth}\centering LuxDiT\end{minipage}%
\begin{minipage}[c]{0.16666\linewidth}\centering GaSLight\end{minipage}%
\begin{minipage}[c]{0.16666\linewidth}\centering Ours\end{minipage}
\end{minipage}\\[1.2pt]
\begin{minipage}[c]{0.018\textwidth}\centering\rotatebox[origin=c]{90}{\scriptsize\textbf{Infinigen}}\end{minipage}%
\hspace{0.012\textwidth}%
\begin{minipage}[c]{0.965\textwidth}\centering
\includegraphics[width=\linewidth]{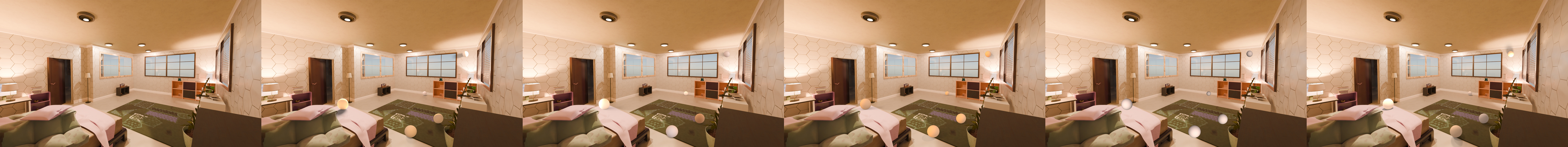}\\[0.7pt]
\includegraphics[width=\linewidth]{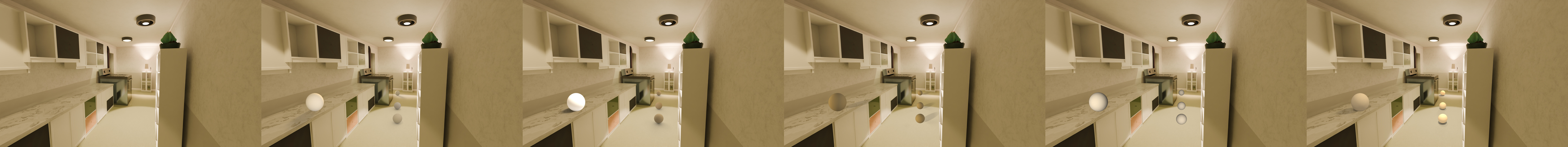}
\end{minipage}\\[1.2pt]
\begin{minipage}[c]{0.018\textwidth}\centering\rotatebox[origin=c]{90}{\scriptsize\textbf{Laval-SV}}\end{minipage}%
\hspace{0.012\textwidth}%
\begin{minipage}[c]{0.965\textwidth}\centering
\includegraphics[width=\linewidth]{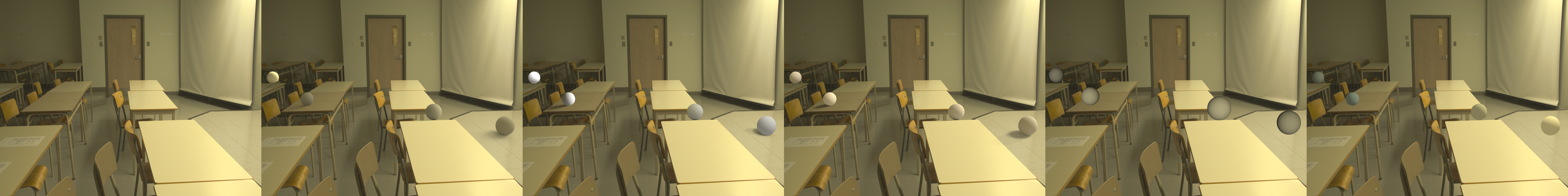}\\[0.7pt]
\includegraphics[width=\linewidth]{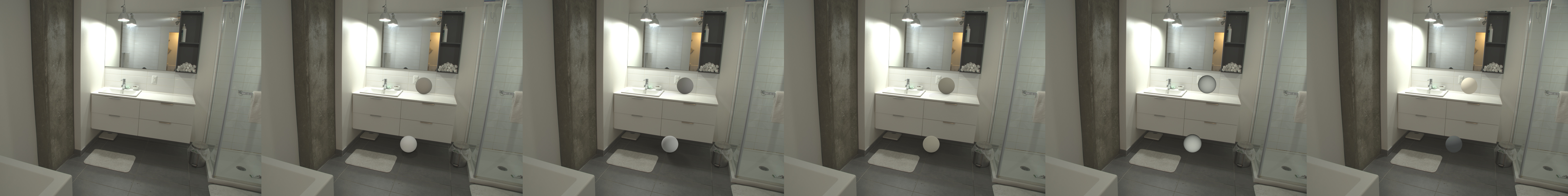}
\end{minipage}\\[1.2pt]
\begin{minipage}[c]{0.018\textwidth}\centering\rotatebox[origin=c]{90}{\scriptsize\textbf{LuxRemix}}\end{minipage}%
\hspace{0.012\textwidth}%
\begin{minipage}[c]{0.965\textwidth}\centering
\includegraphics[width=\linewidth]{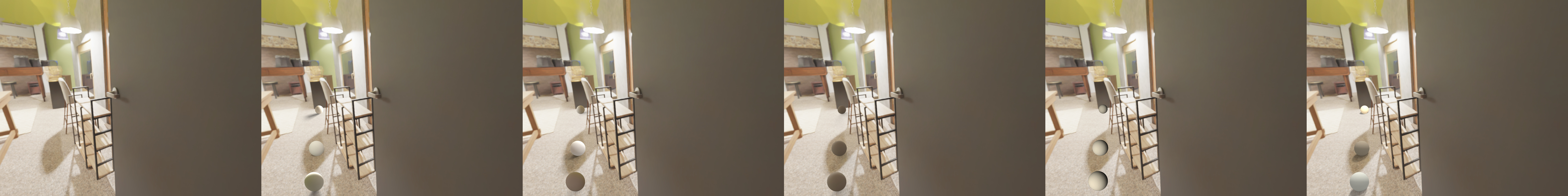}\\[0.7pt]
\includegraphics[width=\linewidth]{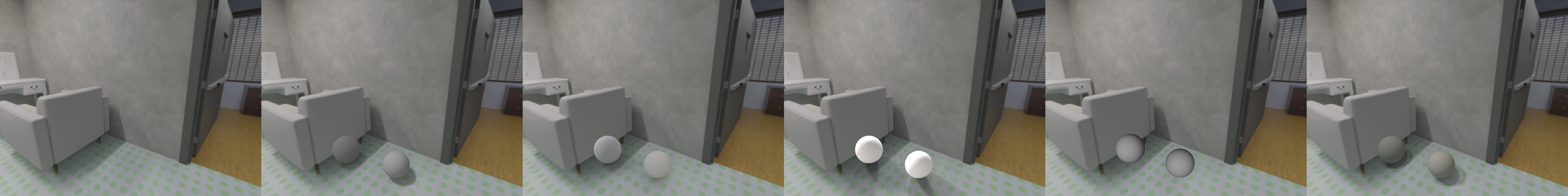}
\end{minipage}\\[1.2pt]
\begin{minipage}[c]{0.018\textwidth}\centering\rotatebox[origin=c]{90}{\scriptsize\textbf{Laval-RF}}\end{minipage}%
\hspace{0.012\textwidth}%
\begin{minipage}[c]{0.965\textwidth}\centering
\includegraphics[width=\linewidth]{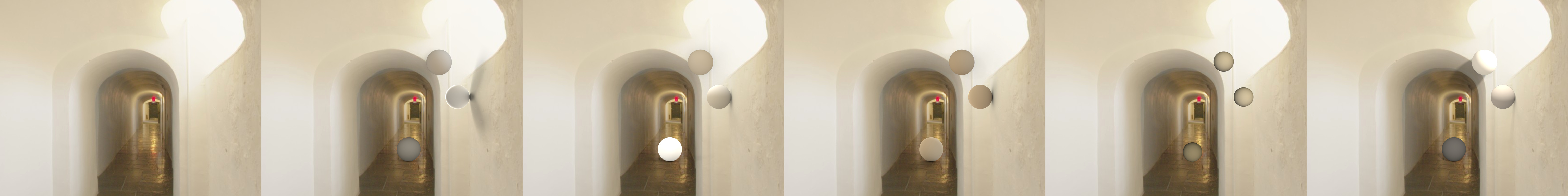}\\[0.7pt]
\includegraphics[width=\linewidth]{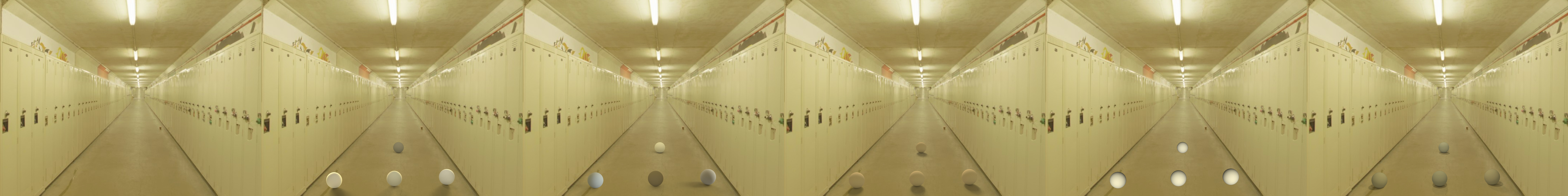}
\end{minipage}\\[1.2pt]
\begin{minipage}[c]{0.018\textwidth}\centering\rotatebox[origin=c]{90}{\scriptsize\textbf{Hypersim}}\end{minipage}%
\hspace{0.012\textwidth}%
\begin{minipage}[c]{0.965\textwidth}\centering
\includegraphics[width=\linewidth]{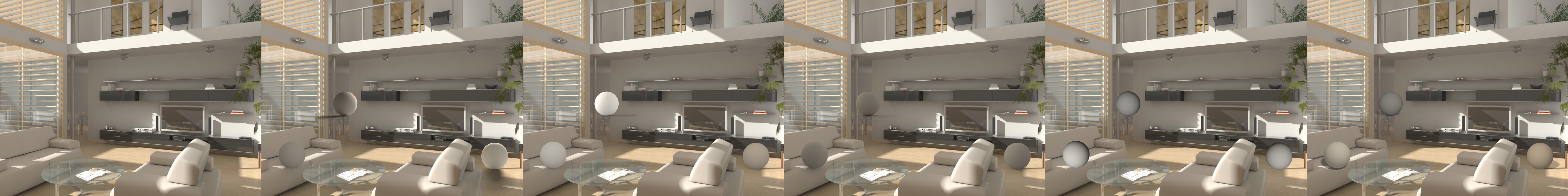}\\[0.7pt]
\includegraphics[width=\linewidth]{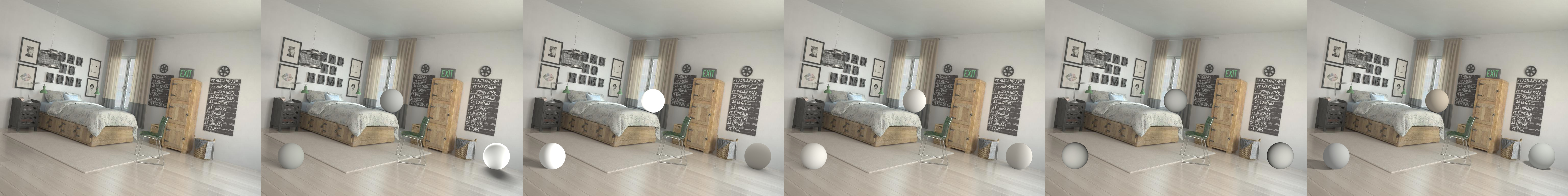}
\end{minipage}
\caption{\textbf{Qualitative probe rendering across datasets.}
Two scenes per dataset and the diffuse spheres are inserted at identical, manually selected query locations for all methods.}
\label{fig:probe_rendering}
\end{figure}

\begin{table}[!ht]
\captionsetup{justification=raggedright,singlelinecheck=false}
\caption{\small{\textbf{Query-to-source direction accuracy on source-aligned Infinigen.}
Our method substantially outperforms lighting-estimation baselines under source-aligned evaluation.}}
\label{tab:direction_aligned_main}
\centering
\small
\setlength{\tabcolsep}{5pt}
\renewcommand{\arraystretch}{1.0}
\begin{tabularx}{\linewidth}{@{}l c >{\centering\arraybackslash}X c >{\centering\arraybackslash}X@{}}
\toprule
& \multicolumn{2}{c}{Physical (\(N_q\)=110)}
& \multicolumn{2}{c}{Aligned \(5^\circ\) (\(N_q\)=59)} \\
\cmidrule(lr){2-3}
\cmidrule(lr){4-5}
Method
& Acc.@$22.5^\circ$ (\%)
& Med. ($^\circ$)
& Acc.@$22.5^\circ$ (\%)
& Med. ($^\circ$) \\
\midrule
IllumiDiff (SG)
& 9.1 & 63.8 & 8.5 & 62.8 \\
DiffusionLight-Turbo (global)
& 2.7 & 80.3 & 1.7 & 90.8 \\
DiffusionLight-Turbo (point)
& 3.6 & 85.4 & 1.7 & 92.3 \\
LiMo
& 26.4 & 51.1 & 32.2 & 45.7 \\
\rowcolor{oursgray}
Ours
& \textbf{86.4}
& \textbf{4.1}
& \textbf{96.6}
& \textbf{3.3} \\
\bottomrule
\end{tabularx}
\end{table}

\paragraph{Direction comparison.}
Previous lighting-estimation baselines predict incident-light distributions, whereas our target is the 3D direction from a query to its dominant physical source. 
Under indirect illumination, even an accurate lighting estimate can yield a direction toward a reflecting surface rather than the physical emitter.
To avoid conflating this target mismatch with prediction error, we use ground-truth renders and source geometry to select Infinigen queries where directions extracted from reference lighting agree with our target within a specified angular tolerance (Appendix~\ref{app:direction_aligned}).
Selection does not use any method's prediction correctness. On these subsets, our model maintains a substantial advantage over the lighting-estimation baselines (Table~\ref{tab:direction_aligned_main}).

\paragraph{Source inventory.}
A natural alternative to VLM-based source localization is to detect visible light fixtures with an object detector and lift them to 3D.
We therefore compare against YOLOv8n fine-tuned on the same training data, using either monocular or ground-truth depth for 3D lifting.
Inventory recovery is evaluated with micro-averaged precision, recall, and F1 using per-image one-to-one matching within distance \(\tau\) (Appendix~\ref{app:inventory_matching}).
Unlike this detector-plus-depth pipeline, our model predicts the 3D source inventory from a single RGB image without an input depth map (Table~\ref{tab:inventory}).


\paragraph{Inference efficiency.}
Efficient lighting estimation is important for applications that require lighting information at many locations within a scene.
Diffusion-based methods such as LiMo and DiffusionLight-Turbo generate light probes separately for each queried location.
Our framework accepts all query points marked together on the input image and produces a shared source inventory and per-query predictions in one autoregressive pass, with a dense irradiance map computed once per image.
This shared computation reduces the cost of estimating lighting at multiple locations.
For 100 query points per scene, our framework achieves the lowest total runtime among the evaluated methods (Figure~\ref{fig:runtime}).

\paragraph{Probe rendering.}
To compare our source-aware representation fairly with previous lighting-estimation methods, we evaluate all methods through the common downstream task of probe rendering.
We additionally evaluate on public external real-world datasets to test generalization beyond our synthetic training domain.
As shown in Table~\ref{tab:radiometry} and Figure~\ref{fig:probe_rendering}, our method remains competitive in rendering quality while additionally recovering explicit source identities and 3D geometry.

\begin{table}[!t]
\centering
\captionsetup{justification=raggedright,singlelinecheck=false}
\caption{\small{\textbf{Active-source inventory localization.}
Our method achieves competitive 3D inventory localization without input depth maps. The fine-tuned detector baseline reaches similar performance only when supplied with ground-truth depth.}}
\label{tab:inventory}
\centering
\small
\setlength{\tabcolsep}{6pt}
\renewcommand{\arraystretch}{1.12}
\begin{tabularx}{\linewidth}{l *{5}{>{\centering\arraybackslash}X}}
\toprule
Method
& F1@.25
& P@.5
& R@.5
& F1@.5
& F1@1.0 \\
\midrule
Qwen3-VL-8B (zero-shot)
& 0.0 & 0.0 & 0.0 & 0.0 & 0.0 \\
YOLOv8n + DAv2 depth
& 0.7 & 2.1 & 1.8 & 2.0 & 18.7 \\
YOLOv8n + GT depth
& 78.4 & 87.9 & 75.1 & 81.0 & 87.5 \\
\rowcolor{oursgray}
Ours
& 52.0 & 76.9 & 76.0 & 76.5 & 87.5 \\
\bottomrule
\end{tabularx}
\end{table}

\begin{table}[!htbp]
\centering
\captionsetup{font=small}
\small
\setlength{\tabcolsep}{4pt}
\renewcommand{\arraystretch}{1.0}
\caption{\textbf{Probe rendering and generalization to external real-world datasets.}
Across Infinigen and public real-world datasets, our method achieves competitive rendering quality while additionally recovering explicit physical sources in 3D.}
\label{tab:radiometry}

\begin{tabular}{@{}llccccc@{}}
\toprule
Dataset
& Method
& LPIPS \(\downarrow\)
& RMSE \(\downarrow\)
& si-RMSE \(\downarrow\)
& SSIM \(\uparrow\)
& RGB Ang.\ (\(^\circ\)) \(\downarrow\) \\
\midrule

\multirow{5}{*}{Infinigen}
& LiMo
& 0.156 & 49.138 & 1.199 & 0.818 & 7.79 \\
& DiffusionLight-Turbo
& 0.198 & 2.518 & 1.487 & 0.748 & 10.60 \\
& LuxDiT
& 0.180 & 2.014 & 1.458 & 0.754 & 7.92 \\
& GaSLight
& 0.280 & 2.605 & 1.467 & 0.684 & 14.17 \\
\oursrowprobe
{Ours}
{0.157}
{1.655}
{0.963}
{0.831}
{6.49}

\midrule

\multirow{6}{*}{Laval-SV}
& ~\citep{Garon_2019_CVPR}
& 0.092 & 1.004 & 0.613 & 0.887 & 6.95 \\
& LiMo
& 0.114 & 1.602 & 0.919 & 0.806 & 4.47 \\
& DiffusionLight-Turbo
& 0.127 & 1.832 & 0.864 & 0.789 & 7.78 \\
& LuxDiT
& 0.119 & 1.626 & 0.889 & 0.814 & 6.36 \\
& GaSLight
& 0.421 & 9.23 & 8.99 & 0.149 & 12.84 \\
\oursrowprobe
{Ours}
{0.114}
{1.642}
{0.602}
{0.849}
{8.67}

\midrule

\multirow{5}{*}{LuxRemix}
& LiMo
& 0.207 & 3.137 & 1.066 & 0.853 & 8.46 \\
& DiffusionLight-Turbo
& 0.206 & 2.051 & 1.202 & 0.855 & 9.79 \\
& LuxDiT
& 0.193 & 1.921 & 0.963 & 0.864 & 8.91 \\
& GaSLight
& 0.435 & 19.770 & 19.558 & 0.197 & 12.07 \\
\oursrowprobe
{Ours}
{0.217}
{2.056}
{0.970}
{0.852}
{11.24}

\midrule

\multirow{5}{*}{Laval-RF}
& LiMo
& 0.155 & 1.329 & 0.489 & 0.928 & 8.80 \\
& DiffusionLight-Turbo
& 0.153 & 1.083 & 0.463 & 0.919 & 7.60 \\
& LuxDiT
& 0.142 & 1.129 & 0.448 & 0.920 & 4.16 \\
& GaSLight
& 0.443 & 13.3 & 13.1 & -- & 10.35 \\
\oursrowprobe
{Ours}
{0.154}
{1.228}
{0.428}
{0.911}
{7.31}

\bottomrule
\end{tabular}
\end{table}

\begin{table}[!t]
\centering
\caption{\small{\textbf{Training-objective ablation.}
CE denotes autoregressive cross-entropy.
Attribution and position denote auxiliary contribution-share supervision and position regression.
Full combines all three objectives.}}
\label{tab:ablation_objective}
\small
\setlength{\tabcolsep}{6pt}
\renewcommand{\arraystretch}{1.12}
\begin{tabularx}{\linewidth}{l *{4}{>{\centering\arraybackslash}X}}
\toprule
\shortstack[l]{Training\\objective}
& \shortstack{PCK@.25\\$\uparrow$}
& \shortstack{PCK@.5\\$\uparrow$}
& \shortstack{PCK@1.0\\$\uparrow$}
& \shortstack{Dir.@$22.5^\circ$\\$\uparrow$} \\
\midrule

CE only
& 47.3 & 69.7 & 78.2 & 83.4 \\

CE + attribution
& 48.3 & 71.3 & 82.8 & 83.3 \\

CE + position
& 45.0 & 68.5 & 79.3 & 83.7 \\

\rowcolor{oursgray}
Full
& 49.2 & 70.6 & 82.2 & 84.6 \\
\bottomrule
\end{tabularx}
\end{table}

\paragraph{Ablation.}
We evaluate the two auxiliary objectives under identical initialization, training data, and training budgets (Table~\ref{tab:ablation_objective}).
Contribution-share supervision improves source localization over cross-entropy alone across all three distance thresholds.
Combining it with auxiliary position regression achieves the highest PCK@0.25\,m and direction accuracy, improving these metrics by 0.9 and 1.3 percentage points over contribution-share supervision alone.
These gains motivate retaining both auxiliary objectives for precise source localization and query-to-source direction prediction.

\section{Conclusion}
We present LAMP, a framework for source-aware indoor illumination estimation that explicitly connects local lighting to the physical sources that produce it. 
A VLM predicts a shared 3D inventory of active sources and attributes each query to contributing sources, while a lightweight network provides dense RGB irradiance. 
Emission-normalized per-source renders enable scalable supervision across lighting configurations. 
LAMP remains competitive with prior lighting-estimation methods in probe-rendering quality on synthetic and real-world data, while additionally providing source identities, 3D positions, and query-to-source relationships.

The main limitation is source depth, which remains the dominant localization error, while secondary-source attribution is less reliable than dominant-source prediction. 
We localize sources rather than recover indirect-light arrival directions, and environment illumination remains unattributed. Future work will improve source depth and recover these directions.
\FloatBarrier

\bibliography{references}
\bibliographystyle{references}
\clearpage
\appendix
\section{Appendix}

\subsection{Additional Method Details}
\subsubsection{Problem Formulation}
\label{app:problem_form}
\paragraph{Inventory-local identifiers.}
\label{app:problem_inventory_identifier}
Each active source is assigned an identifier (\texttt{L1}, \texttt{L2}, etc.) local to the current example. 
To construct the target inventory, we order in-frame sources from left to right by their projected horizontal coordinates, followed by off-screen sources ordered by their azimuth in the camera frame (measured about the camera's up axis from the viewing direction).
A source counts as in-frame when the renderer marks its fixture inside the field of view and the projection of its emitter position falls inside the image. 
The azimuth is \(\operatorname{atan2}(x,-z)\) of the emitter position in the camera frame, in \((-\pi,\pi]\), and ties in either ordering are broken by the order of the sources in the scene file. 
This deterministic ordering eliminates arbitrary label permutations and maintains a consistent correspondence between the inventory and the query-conditioned output within each training example.
Physical emitter identities are preserved across matched configurations, whereas inventory-local identifiers are assigned independently within each example and may change when the active source set changes. 

\paragraph{Query-conditioned output.}
\label{app:problem_query_output}
The 2D direction \(\widehat{\boldsymbol{\omega}}_{i,r}^{\mathrm{2D}}\in\mathbb{R}^{2}\) is the normalized image-space displacement from the query marker to the projection of the source emitter position, computed on the focal-length-normalized image. 
For sources whose inventory pixel is \texttt{null}, that is, sources projecting outside the image or behind the camera, the target is \([0,0]\) and the entry is excluded from 2D evaluation. 
The 2D direction is used only for 2D evaluation.

\subsubsection{Per-Light Transport and Controlled Configurations}
\label{app:emission_transport_basis}

\paragraph{Emission and reference renders.}
The effective emission of source \(s_k\) is \(\boldsymbol{\eta}_{s_k}^{l}=E_{s_k}^{l}\,\mathbf{c}_{s_k}^{l}\), the product of the renderer's light power parameter \(E_{s_k}^{l}\) (in watts) and the RGB emission color \(\mathbf{c}_{s_k}^{l}\). 
The reference emission used for the solo render is non-zero in every channel, so the channel-wise division in~\eqref{eq:emission_normalized_contribution} is well defined. 

\paragraph{Matched configurations.}
Multipliers \(\boldsymbol{\alpha}_{s_k}^{l}\in\mathbb{R}^{3}_{\geq 0}\) act channel-wise, so an intervention rescales source power and may change its emission color while 
 the emitter's spatial emission profile, scene geometry, materials, camera, and source identities remain fixed. We include source-strengthening and source-weakening interventions, and configurations in which the dominant source of a region changes as well as ones in which it stays fixed.
Each configuration receives its own structured and share supervision.

\paragraph{Share computation.}
\(\operatorname{lum}\) uses the Rec.~709 weights \((0.2126,0.7152,0.0722)\). 
Negative differences between the solo render and the baseline, which arise from Monte Carlo noise, are clipped to zero per channel before the luminance conversion, and queries whose total source luminance \(\sum_{j=1}^{K^{l}}\operatorname{lum}(\mathbf{C}_{s_j}^{l}(\mathbf{q}_i))\) is zero are discarded. 
The evaluation annotations retain the same truncated lists for the all-slot metrics.

\begin{figure}[H]
    \centering
    \includegraphics[width=\linewidth]{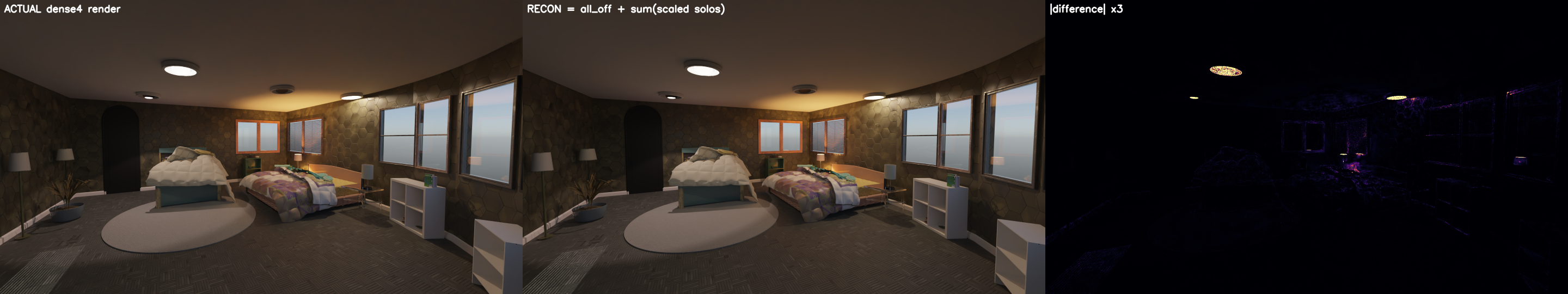}
    \caption{
    \textbf{Render-level validation of emission-normalized superposition.}
    The left panel shows a composite generated directly by the renderer, the middle panel its reconstruction using 
    \(\mathbf{L}_{\mathrm{off}}+\sum_{k=1}^{K^{l}}\mathbf{C}_{s_k}^{l}\), 
    and the right panel the absolute pixel-wise difference, amplified by \(3\times\) for
    visibility. The reconstructed and directly rendered composites closely
    agree in linear RGB radiance space.
    }
    \label{fig:additivity_validation}
\end{figure}

\paragraph{Composite reconstruction and additivity validation.} 
We compare the reconstruction \(\widehat{\mathbf{L}}^{l}(\mathbf{u})=\mathbf{L}_{\mathrm{off}}(\mathbf{u})+\sum_{k=1}^{K^{l}}\mathbf{C}_{s_k}^{l}(\mathbf{u})\) with the directly rendered composite of configuration \(l\) in linear RGB radiance space. 
The residual of a configuration is the mean absolute difference in luminance between the two images divided by the mean luminance of the rendered composite, with both images clipped at their 99.9th percentile to suppress fireflies. 
The median residual over configurations is 3.5\% on Infinigen and 3.0\% on LuxRemix, supporting the accuracy of emission-normalized superposition under our rendering setup. 
Figure~\ref{fig:additivity_validation} shows a qualitative comparison, and configurations are not filtered using this residual. 

\subsubsection{Inference and Count-Consistent Decoding}
\label{app:constrained_decoding}
Given only the marked image and prompt, the model generates the inventory and query records by greedy autoregressive decoding, without training-time renders or annotations.
If a query's list contains fewer than \(\min(\widehat{n}_i,N_D)\) entries, we resume from its last entry and use a regular-expression constraint to append complete entries referencing unused inventory identifiers until the target length is reached.
Completion only appends missing entries with unused inventory identifiers, leaving all existing predictions unchanged.
For the final model, every greedy list already matches its declared count up to \(N_D\) on both held-out splits, so completion is never triggered and all reported results use a single greedy pass.

\subsubsection{Dense Irradiance Map Details}
\label{app:irradiance_details}

\paragraph{Architecture and training.}
The dense network is a ResNet-18 encoder with a feature-pyramid decoder.
It takes the unmarked LDR image at \(640\times360\) and outputs a \(160\times90\) map of log RGB irradiance, one quarter of the input resolution.
It is trained only on the Infinigen views and configurations underlying the structured training images, re-rendered with the fixture-color augmentation below; LuxRemix has no irradiance targets. Training uses a per-pixel smooth-L1 loss on log irradiance plus a gradient term, and inference runs once per image.

\paragraph{Target derivation.}
For a Lambertian surface with albedo \(\rho\), the outgoing diffuse radiance is \((\rho/\pi)\,\mathbf{E}\), where \(\mathbf{E}\) is the irradiance arriving at the surface~\citep{kajiya1986rendering}.
The renderer's diffuse light passes report the direct and indirect diffuse radiance with the albedo divided out according to Blender cycles manual, so their sum equals \(\mathbf{E}/\pi\).
We therefore take
\begin{equation}
\mathbf{E}(\mathbf{u})=\pi\,\bigl(\mathbf{L}_{\mathrm{dir}}(\mathbf{u})+\mathbf{L}_{\mathrm{ind}}(\mathbf{u})\bigr)
\label{eq:irradiance_target}
\end{equation}
as the irradiance at the surface seen through pixel \(\mathbf{u}\), where \(\mathbf{L}_{\mathrm{dir}}\) and \(\mathbf{L}_{\mathrm{ind}}\) are these albedo-free diffuse passes.
Pixels whose diffuse albedo is negligible, such as emitters and purely specular surfaces, are masked out of the target.
Unlike the per-source contributions of Section~\ref{sec:per_light_transport}, which are radiance seen by the camera, this target is the light arriving at the surface, summed over all sources and the environment.

\paragraph{Fixture-color augmentation.}
Every Infinigen fixture emits the same default color, so a map trained on the original renders could infer chromaticity from fixture color alone.
We re-render each training scene with a blackbody color temperature drawn per fixture from \([3500, 9000]\,\)K and desaturated by 20\%, fixed per scene, and export the inputs through the same display transform and exposure as the dataset images.

\paragraph{Calibration to the probe protocol.}
The map predicts irradiance at a surface, whereas the probe protocol of Section~\ref{sec:experiment_results} measures the mean radiance \(\bar{L}_{\mathrm{probe}}\) of a white Lambertian sphere inserted at the query, which collects light from directions the surface does not see and is shadowed by its own body.
The regressor also compresses its predictions toward the mean.
We correct both effects with one affine map in log space,
\begin{equation}
\log \bar{L}_{\mathrm{probe}} = a\,\log E_{\mathrm{map}} + b,
\label{eq:probe_calibration}
\end{equation}
where \(E_{\mathrm{map}}\) is the luminance of the map at the query pixel.
The two parameters are fitted once on probes rendered under the true lighting at calibration queries in the training scenes and are shared by all test images, so no ground truth is used at test time.

\paragraph{Probe composition from the predicted outputs.}
\label{app:probe_composition}
This section describes the downstream use evaluated in Section~\ref{sec:experiment_results}.
The two branches are complementary. The map supplies brightness, color, and their variation across the image, which the VLM does not read out reliably, and the ranked sources supply the direction of shading, the cast shadow, and the highlight, which the map cannot provide because its spatial gradient is not a usable direction cue.
For query \(\mathbf{q}_i\) the VLM provides the ranked sources \(\{(\widehat{s}_{i,r}, \widehat{\boldsymbol{\omega}}_{i,r}, \widehat{w}_{i,r})\}_{r=1}^{\widehat m_i}\) and the map provides the calibrated luminance \(\ell_i\) and chromaticity \(\mathbf{c}_i\).
We place one virtual emitter per ranked entry along its predicted direction \(\widehat{\boldsymbol{\omega}}_{i,r}\), using the distance from the query to the referenced inventory position (at least 0.5\,m; 3\,m if unlocalized), with power proportional to its share \(\widehat{w}_{i,r}\).
The more accurate direction prediction sets the emitter direction; the inventory position supplies only its distance.
An isotropic fill term accounts for the light not attributed to any emitter.
We assign a fixed fraction \(f_{\mathrm{fill}}=0.65\) of the target luminance to the fill and the remainder to the emitters, and scale both so that the analytic mean radiance of a white Lambertian sphere at the query equals \(\ell_i\) with chromaticity \(\mathbf{c}_i\).
A probe sphere is then rendered under these emitters with a physically based renderer, which yields the cast shadow and, for a glossy probe, the highlight that a direction-free estimate cannot produce.
Only the query's own geometry, its 3D position and surface normal, enters this step.
For photographs, we estimate metric depth with Depth Anything V2 (metric indoor, large), bicubically upsample the prediction to the original image resolution, and take the median depth in a $5\times5$ window centered at the query. The query position is then back-projected using the pinhole intrinsics of the image crop. To estimate its surface normal, we back-project a $21\times21$ neighborhood around the query, fit a plane by least squares, and orient the resulting normal toward the camera. Before rendering photographs, the predicted chromaticity $\mathbf{c}_i$ is additionally corrected with gray-world white balancing, consistent with Table~\ref{tab:radiometry}.
For Laval-SV, the probe is placed at the measured chrome-ball location. Its depth is recovered from the ball's known physical radius and observed pixel radius, and the surface normal is set to the vertical direction. For Hypersim and Infinigen, query depth and normals come from the renderer. For the additional qualitative query locations where a renderer normal is not directly stored, we apply the same local plane-fitting procedure to the rendered depth map.

\subsection{Experiment Implementation Details}
Training uses an Infinigen single-marker initialization stage followed by main-stage structured training on the full training set.
The initialization stage fine-tunes Qwen3-VL-8B-Instruct for five epochs on 14{,}147 single-marker direction examples from 84 Infinigen training scenes, and the main stage then fine-tunes this checkpoint for four epochs on the structured objective of Section~\ref{sec:method} over the full training set of Section~\ref{sec:experimental_setup}.
Both stages use AdamW with a peak learning rate of \(3\times10^{-6}\), a cosine schedule with 10\% warm-up, weight decay 0.01, a global batch size of 4, bfloat16, and FSDP on four A100 80\,GB GPUs, with \(\lambda_{\mathrm{attr}}=0.5\) and \(\lambda_{\mathrm{reg}}=0.1\) fixed a priori and not tuned.
The number of epochs was fixed in advance and we evaluate the final checkpoint. 
The auxiliary objectives \(\mathcal{L}_{\mathrm{attr}}\) and \(\mathcal{L}_{\mathrm{reg}}\) are used in every trained variant except the loss ablations that remove them, and their heads are discarded after training.
The dense irradiance network is trained on the Infinigen training views only, using the 1{,}989 unique renders of the 111 training scenes that underlie the marked training images, re-rendered with the per-fixture color-temperature augmentation of Appendix~\ref{app:irradiance_details}, with the same 10 validation scenes (211 configurations) for model selection.
It takes the unmarked image at \(640\times360\) and predicts a \(160\times90\) log-irradiance map with a ResNet-18 feature-pyramid decoder, trained for 6{,}000 steps with AdamW (learning rate \(3\times10^{-4}\), weight decay \(10^{-4}\), cosine schedule, batch size 16) on a smooth-L1 loss over log irradiance plus a gradient term, selecting the seed and step with the lowest validation log error.
Its calibration to the probe protocol is fitted only on probes rendered in the Infinigen training scenes, and no test-set probe reference or photograph is used for training or calibration.

\subsubsection{Inventory Matching Protocol}
\label{app:inventory_matching}
For each image, predicted and ground-truth sources are matched by a maximum-cardinality matching under the distance constraint, computed with the Hungarian algorithm on a cost matrix that holds the camera-frame 3D distance for pairs within \(\tau\) and a prohibitive constant otherwise, which maximizes the number of matched pairs and, among such matchings, minimizes their total distance. 
Each ground-truth source is matched at most once, so repeated predictions of the same source count as false positives. 
True positives, unmatched ground-truth sources, and unmatched predictions are summed over the 168 images of the Infinigen test set before precision, recall, and F1 are formed. 
A greedy variant that processes ground-truth sources in inventory order and assigns each the nearest unmatched prediction gives identical numbers on every row of Table~\ref{tab:inventory}. 

\subsection{Additional Experiment Results}
\subsubsection{Source Localization and Direction Results}
\label{app:single_multi}
\paragraph{Full dominant-source results.}
Fine-tuning substantially improves dominant-source localization over the zero-shot model on both test sets. On Infinigen, the model localizes 70.6\% of dominant sources within 0.5\,m, and among these correctly localized cases, 98.2\% of predicted directions fall within \(22.5^\circ\). Localization is more challenging on LuxRemix, while the conditional direction estimates remain substantially more accurate than the zero-shot baseline.
\begin{table}[H]
\centering
\scriptsize
\setlength{\tabcolsep}{2.6pt}
\renewcommand{\arraystretch}{1.08}
\caption{\textbf{Full dominant-source results.}
Conditional direction metrics are restricted to queries whose dominant source is localized within 0.5\,m; the all-query block scores every query, counting a missing or unparsable direction as incorrect, with medians over valid outputs, and Valid is the share of queries whose top-ranked entry carries a parsable 3D direction.
\textbf{+ GT depth} retains predicted source pixels and replaces their depth with ground truth, which leaves the predicted directions unchanged. Dashes denote undefined results.}
\label{tab:source_referenced_full}
\begin{tabular}{@{}l ccc cccc ccccc@{}}
\toprule
& \multicolumn{3}{c}{3D PCK (\%) $\uparrow$}
& \multicolumn{4}{c}{Direction $\mid$ PCK@0.5\,m}
& \multicolumn{5}{c}{Direction, all queries} \\
\cmidrule(lr){2-4}\cmidrule(lr){5-8}\cmidrule(lr){9-13}
Model
& 0.25\,m & 0.5\,m & 1.0\,m
& \multicolumn{3}{c}{Accuracy (\%) $\uparrow$}
& Med. ($^\circ$) $\downarrow$
& \multicolumn{3}{c}{Accuracy (\%) $\uparrow$}
& Med. ($^\circ$) $\downarrow$
& Valid (\%) \\
\cmidrule(lr){5-7}\cmidrule(lr){9-11}
& & & & $12.5^\circ$ & $22.5^\circ$ & $30^\circ$ & & $12.5^\circ$ & $22.5^\circ$ & $30^\circ$ & & \\
\midrule
\multicolumn{13}{@{}l}{\textsc{Infinigen (all 1{,}032 test queries)}} \\
Qwen3-VL-8B (zero-shot)
& 0.0 & 0.0 & 0.0 & -- & -- & -- & -- & 2.4 & 7.2 & 10.3 & 99.4 & 91.7 \\
\quad + GT depth
& 0.7 & 1.6 & 5.0 & 0.0 & 0.0 & 0.0 & 85.0 & 2.4 & 7.2 & 10.3 & 99.4 & 91.7 \\
Qwen3-VL-32B (zero-shot)
& 0.0 & 0.0 & 0.0 & -- & -- & -- & -- & 0.4 & 2.1 & 3.4 & 107.5 & 90.8 \\
\quad + GT depth
& 1.0 & 1.4 & 5.4 & 0.0 & 0.0 & 0.0 & 90.3 & 0.4 & 2.1 & 3.4 & 107.5 & 90.8 \\
\rowcolor{oursgray}
Ours
& 49.2 & 70.6 & 82.2 & 94.5 & 98.2 & 98.8 & 2.7 & 78.3 & 84.6 & 87.6 & 3.9 & 100.0 \\ 
\midrule
\multicolumn{13}{@{}l}{\textsc{LuxRemix (all 9{,}412 test queries)}} \\
Qwen3-VL-8B (zero-shot)
& 0.0 & 0.1 & 0.4 & -- & -- & -- & -- & 0.5 & 1.9 & 3.5 & 92.9 & 80.0 \\
\quad + GT depth
& 2.4 & 6.0 & 19.3 & 0.7 & 2.8 & 3.9 & 90.0 & 0.5 & 1.9 & 3.5 & 92.9 & 80.0 \\
\rowcolor{oursgray}
Ours
& 10.1 & 28.5 & 46.0 & 65.7 & 87.6 & 93.9 & 9.1 & 32.1 & 50.2 & 59.6 & 22.3 & 99.8 \\ 
\bottomrule
\end{tabular}
\end{table}
\paragraph{Performance with different numbers of active sources.}
The main drop from single- to multi-source scenes is in source localization: PCK@0.5\,m decreases from 82.5\% to 64.6\%. In contrast, once the dominant source is localized within 0.5\,m, direction prediction remains stable, with 99.0\% and 97.7\% of directions within \(22.5^\circ\) for single- and multi-source scenes, respectively.
\begin{table}[H]
\centering
\scriptsize
\setlength{\tabcolsep}{2.6pt}
\renewcommand{\arraystretch}{1.08}
\caption{\textbf{Source-referenced results by number of active sources (Infinigen test set).} Metrics as in Table~\ref{tab:main_source_referenced}; the all-query direction block and Valid are defined as in Table~\ref{tab:source_referenced_full}.}
\label{tab:single_multi}
\begin{tabular}{@{}l ccc ccc c ccc c c@{}}
\toprule
& \multicolumn{3}{c}{3D PCK (\%) \(\uparrow\)} & \multicolumn{3}{c}{Dir \(\mid\) PCK@0.5\,m (\%) \(\uparrow\)} & Med. & \multicolumn{3}{c}{Dir, all queries (\%) \(\uparrow\)} & Med. & Valid \\
\cmidrule(lr){2-4}\cmidrule(lr){5-7}\cmidrule(lr){8-8}\cmidrule(lr){9-11}\cmidrule(lr){12-12}\cmidrule(lr){13-13}
Model & 0.25\,m & 0.5\,m & 1.0\,m & \(12.5^\circ\) & \(22.5^\circ\) & \(30^\circ\) & (\(^\circ\)) \(\downarrow\) & \(12.5^\circ\) & \(22.5^\circ\) & \(30^\circ\) & (\(^\circ\)) \(\downarrow\) & (\%) \\
\midrule
\multicolumn{13}{@{}l}{\textsc{Single Source} (348 queries, 59 images)} \\
Qwen3-VL-8B (zero-shot) & 0.0 & 0.0 & 0.0 & -- & -- & -- & -- & 3.4 & 8.0 & 10.9 & 98.4 & 86.8 \\
Qwen3-VL-32B (zero-shot) & 0.0 & 0.0 & 0.0 & -- & -- & -- & -- & 0.6 & 2.0 & 3.2 & 108.5 & 91.4 \\
\rowcolor{oursgray} Ours & 52.6 & 82.5 & 94.0 & 96.5 & 99.0 & 99.7 & 2.4 & 93.1 & 96.3 & 97.4 & 2.6 & 100.0 \\ 
\midrule
\multicolumn{13}{@{}l}{\textsc{Multi Sources} (684 queries, 109 images)} \\
Qwen3-VL-8B (zero-shot) & 0.0 & 0.0 & 0.0 & -- & -- & -- & -- & 1.9 & 6.7 & 9.9 & 99.7 & 94.2 \\
Qwen3-VL-32B (zero-shot) & 0.0 & 0.0 & 0.0 & -- & -- & -- & -- & 0.3 & 2.2 & 3.5 & 107.3 & 90.5 \\
\rowcolor{oursgray} Ours & 47.5 & 64.6 & 76.2 & 93.2 & 97.7 & 98.2 & 3.0 & 70.8 & 78.7 & 82.6 & 5.0 & 100.0 \\ 
\bottomrule
\end{tabular}
\end{table}

\paragraph{Dominant source in-frame vs out-of-frame.}
Dominant-source localization is substantially easier when the source is visible in the image. At PCK@0.5\,m, accuracy drops from 73.9\% to 14.3\% for off-frame sources on Infinigen and from 42.0\% to 10.2\% on LuxRemix, indicating that off-frame source localization remains a major failure mode.
\begin{table}[H]
\centering
\small
\setlength{\tabcolsep}{4pt}
\renewcommand{\arraystretch}{1.05}
\caption{\textbf{Dominant-source localization by visibility of the source (Ours).} Queries are split by whether the ground-truth dominant source lies inside the image frame. 3D PCK@\(\tau\) as in Table~\ref{tab:main_source_referenced}.}
\label{tab:inframe_offframe}
\begin{tabular}{@{}l l r ccc@{}}
\toprule
Set & Dominant source & \(N\) (\%) & \(\tau\)=0.25\,m & \(\tau\)=0.5\,m & \(\tau\)=1.0\,m \\
\midrule
Infinigen & in frame & 976 (94.6) & 51.7 & 73.9 & 84.9 \\
Infinigen & off frame & 56 (5.4) & 5.4 & 14.3 & 33.9 \\
LuxRemix & in frame & 5{,}412 (57.5) & 15.9 & 42.0 & 60.6 \\
LuxRemix & off frame & 4{,}000 (42.5) & 2.2 & 10.2 & 26.3 \\
\bottomrule
\end{tabular}
\end{table}

\subsubsection{3D Query-to-Source Direction on a Ground-Truth-Verified Subset}
\label{app:direction_aligned}
Lighting-estimation baselines output lighting distributions from which we extract a direction for comparison with the 3D query-to-source target.
These directions may differ under indirect illumination.
We identify queries where directions extracted from a reference built from ground-truth renders and source geometry agree with the target within a specified tolerance.
Selection uses no prediction correctness, and its rules were fixed before scoring the resulting subsets, after the full-set and direct-stratum analyses.

\paragraph{Physical filtering.}
Using per-source direct and indirect diffuse passes, we require the dominant source to contribute at least 80\% of its own total through direct paths, its direct contribution to account for at least 60\% of total active-source irradiance, and its total contribution to exceed each other source's contribution by a factor of at least two.
Inspection of the references revealed queries on fixtures or thin and translucent surfaces with positive direct passes despite the source lying behind the visible surface.
Before subset scoring, we therefore added three conditions requiring the dominant source to lie within \(80^\circ\) of the surface normal, at least 0.5\,m from the query, and to provide direct irradiance of at least 0.02, the fifth percentile over the candidates.
Together, these conditions retain 110 of 1{,}032 queries from 41 scenes.
References were rendered for 185 candidates from 56 scenes under looser relative thresholds of 70\%, 50\%, and 1.5, of which 161 from 52 scenes satisfy the added conditions.
This broader pool supplies supplementary diagnostics only, and aligned subsets are drawn from the 110-query set.

\paragraph{Local lighting reference.}
We render a 512-pixel-wide equirectangular panorama in Cycles under the input image's lighting, offsetting the query by 1\,cm along the surface normal to avoid self-intersection.
Across the candidates, the offset changes the direction to the dominant source by a median of \(0.14^\circ\) and at most \(1.5^\circ\), compared with \(1.9^\circ\) and \(18.6^\circ\) for the rendering protocol's 0.13\,m probe-centre offset.
Emissive markers along the camera axes verify the panorama's pixel-to-direction mapping.

Camera rays do not directly see the scene's finite-radius point lights, so we supplement the panorama's indirect and environment illumination with approximate direct contributions.
Each source with positive shadowed direct irradiance \(E_k\) at the original query receives a uniform disk centred on its direction from the panorama camera.
The disk has angular radius \(\arctan(r_k/d_k)\), with a three-pixel minimum, where \(r_k\) is the source radius and \(d_k\) its distance from the panorama camera.
Its radiance is \(L_k=E_k/(\Omega_k\cos\theta_k)\), where \(\Omega_k\) is its solid angle and \(\theta_k\) the incidence angle of its centre.
The implementation clamps \(\cos\theta_k\) at 0.05 for secondary sources near the tangent plane.
This clamp never applies to the dominant source under the \(80^\circ\) condition.
The construction approximately preserves direct irradiance using a constant-cosine approximation, with visibility and attenuation taken from the original query.
It does not reconstruct within-disk variations from partial occlusion or changes in visibility at the offset position.

\paragraph{Direction extraction and verification.}
We convert each reference to the representation of each baseline and use the same extraction procedure as for its predictions.
These are the map peak region for DiffusionLight-Turbo, a 12-lobe SG fit followed by the lobe extraction rule for IllumiDiff, a mirror ball synthesized with LiMo's geometry followed by its chrome-ball extraction rule, order-3 SH projection followed by the \(l=1\) direction for UniLight~\citep{zhang2026unilight}, and 96-anchor discretization followed by cluster extraction for EMLight~\citep{zhan2021emlight}.
Table~\ref{tab:gt_input_check} compares these extracted directions with the original 3D query-to-source direction.
The results characterize reference conversion and direction extraction jointly, rather than the representations' intrinsic capacity.

\paragraph{Aligned subsets.}
We retain queries where the map, SG, and chrome-ball estimates from the reference all match the query-to-source direction within \(\delta\).
This yields 59 queries from 29 scenes for \(\delta=5^\circ\), and 60 from 29 scenes for \(10^\circ\).
Table~\ref{tab:direction_aligned} re-scores existing predictions without re-inference.
EMLight and UniLight are reported on the same queries but do not participate in the alignment criterion.
For global DiffusionLight-Turbo, the check validates extraction from a local reference, not the adequacy of one lighting estimate for every query.
These selected diagnostics reduce target mismatch but retain differences in training domain, input, and spatial grounding, and complement the full-set evaluation.

\paragraph{Sensitivity to disk width.}
Typical source radii of 2.4\,cm subtend about \(0.5^\circ\) at 2.7\,m, below the map resolution.
For disk-radius floors of 1, 3, and 6 pixels (approximately \(0.7^\circ\), \(2.1^\circ\), and \(4.2^\circ\)), reference-based direction accuracy within \(5^\circ\) ranges from 91--93\% for the map and 70--77\% for the chrome ball over the 110 queries.
SG accuracy is 43\%, 69\%, and 82\%, yielding aligned subsets of 37, 59, and 73 queries, respectively.
Across these subsets, ours achieves approximately 92--97\% query-to-source direction accuracy within \(22.5^\circ\), compared with 31--32\% for LiMo and at most 15\% for the other methods.
Subset membership therefore depends on reference discretization, while the performance gap persists.
All tables use the three-pixel floor.

\begin{table}[t]
\centering
\small
\setlength{\tabcolsep}{4pt}
\renewcommand{\arraystretch}{1.05}
\caption{\textbf{3D query-to-source direction extraction from ground-truth lighting.}
Errors are relative to the original query-to-source target on 110 queries from 41 scenes.
Parentheses show results on the broader 161-query pool from 52 scenes.
Median errors are in degrees and angular accuracies in percent.}
\label{tab:gt_input_check}
\begin{tabular}{@{}l ccc@{}}
\toprule
Representation & Median (\(^\circ\)) & \(\le 5^\circ\) (\%) & \(\le 10^\circ\) (\%) \\
\midrule
Environment map (DiffusionLight-Turbo) & 0.3 (0.3) & 92 (94) & 93 (95) \\
12 spherical Gaussians (IllumiDiff) & 1.2 (1.2) & 69 (68) & 70 (68) \\
Mirror ball (LiMo) & 0.5 (0.5) & 77 (78) & 78 (78) \\
Order-3 spherical harmonics (UniLight~\citep{zhang2026unilight}) & 9.1 (9.4) & 23 (20) & 55 (53) \\
96 sphere anchors (EMLight~\cite{zhan2021emlight}) & 8.2 (8.0) & 17 (18) & 65 (67) \\
\bottomrule
\end{tabular}
\end{table}

\begin{table}[t]
\centering
\small
\setlength{\tabcolsep}{3pt}
\renewcommand{\arraystretch}{1.05}
\caption{\textbf{3D query-to-source direction on ground-truth-verified Infinigen subsets.}
Physical includes 110 queries from 41 scenes.
Aligned \(5^\circ\) and \(10^\circ\) include 59 and 60 queries, respectively, each from 29 scenes.
Angular accuracies are in percent (higher is better), and median errors in degrees (lower is better).}
\label{tab:direction_aligned}
\begin{tabularx}{\linewidth}{@{}p{0.27\linewidth} *{9}{>{\centering\arraybackslash}X}@{}}
\toprule
& \multicolumn{3}{c}{Physical (110)} & \multicolumn{3}{c}{Aligned \(5^\circ\) (59)} & \multicolumn{3}{c}{Aligned \(10^\circ\) (60)} \\
\cmidrule(lr){2-4}\cmidrule(lr){5-7}\cmidrule(lr){8-10}
Model & \shortstack{Acc.@\\$12.5^\circ$\\(\%)} & \shortstack{Acc.@\\$22.5^\circ$\\(\%)} & \shortstack{Med.\\($^\circ$)} & \shortstack{Acc.@\\$12.5^\circ$\\(\%)} & \shortstack{Acc.@\\$22.5^\circ$\\(\%)} & \shortstack{Med.\\($^\circ$)} & \shortstack{Acc.@\\$12.5^\circ$\\(\%)} & \shortstack{Acc.@\\$22.5^\circ$\\(\%)} & \shortstack{Med.\\($^\circ$)} \\
\midrule
\multicolumn{10}{@{}l}{\textit{Methods included in subset verification}} \\
IllumiDiff (SG) & 3.6 & 9.1 & 63.8 & 1.7 & 8.5 & 62.8 & 1.7 & 8.3 & 62.6 \\
DiffusionLight-Turbo (global reference) & 1.8 & 2.7 & 80.3 & 1.7 & 1.7 & 90.8 & 1.7 & 1.7 & 90.6 \\
DiffusionLight-Turbo (point) & 0.9 & 3.6 & 85.4 & 1.7 & 1.7 & 92.3 & 1.7 & 1.7 & 92.3 \\
LiMo & 10.9 & 26.4 & 51.1 & 13.6 & 32.2 & 45.7 & 13.3 & 33.3 & 45.5 \\
\midrule
\multicolumn{10}{@{}l}{\textit{Other methods}} \\
EMLight & 6.4 & 10.9 & 69.5 & 3.4 & 10.2 & 65.0 & 3.3 & 10.0 & 65.6 \\
UniLight & 6.4 & 13.6 & 57.7 & 5.1 & 13.6 & 53.5 & 5.0 & 13.3 & 53.3 \\
\midrule
\rowcolor{oursgray} Ours & 78.2 & 86.4 & 4.1 & 88.1 & 96.6 & 3.3 & 86.7 & 96.7 & 3.4 \\
\bottomrule
\end{tabularx}
\end{table}

\subsubsection{Additional Source-Inventory Results}
\paragraph{Off-frame sources in the inventory.}
\label{app:inventory_offframe}
Of the 329 ground-truth sources of the Infinigen test set, 24 lie outside the image frame. Table~\ref{tab:inventory_offframe} reports the recall of these sources under the matching of Table~\ref{tab:inventory}; the detector baselines can only recover an off-frame source through a matched in-frame prediction that happens to fall within \(\tau\).
\begin{table}[H]
\centering
\small
\setlength{\tabcolsep}{4pt}
\renewcommand{\arraystretch}{1.05}
\caption{\textbf{Recall of off-frame ground-truth sources (Infinigen test set, 24 of 329 sources).} Same matching as Table~\ref{tab:inventory}; counts of recovered sources in parentheses.}
\label{tab:inventory_offframe}
\begin{tabular}{@{}l ccc@{}}
\toprule
Method & R\(_{\text{off}}\)@0.25\,m \(\uparrow\) & R\(_{\text{off}}\)@0.5\,m \(\uparrow\) & R\(_{\text{off}}\)@1.0\,m \(\uparrow\)\\
\midrule
Qwen3-VL-8B (zero-shot) & 0.0 (0/24) & 0.0 (0/24) & 0.0 (0/24) \\
YOLOv8n (fine-tuned) + GT depth & 4.2 (1/24) & 4.2 (1/24) & 4.2 (1/24) \\
YOLOv8n (fine-tuned) + DA depth & 0.0 (0/24) & 0.0 (0/24) & 0.0 (0/24) \\
\rowcolor{oursgray} Ours & 4.2 (1/24) & 12.5 (3/24) & 25.0 (6/24) \\
\bottomrule
\end{tabular}
\end{table}

\subsubsection{Intervention Benchmark}
\label{app:intervention}
Each sequence fixes a scene, a camera, and a query point that receives light from two sources A and B, and re-renders the view while the emission of B is scaled in steps relative to A, so that the dominant source of the query switches from A to B at a known step. 
Sequences are generated from 70 held-out scenes and audited so that both sources contribute to the query and the crossing lies inside the sequence, leaving 114 sequences with 1{,}078 frames. 
Each frame is decoded independently and the metrics of Table~\ref{tab:intervention} aggregate per sequence. 
The attribution-head metrics are diagnostic only: they are computed from the trained auxiliary head before it is discarded and are not used for decoding or inference.
\begin{table}[H]
\centering
\small
\setlength{\tabcolsep}{4pt}
\renewcommand{\arraystretch}{1.05}
\caption{\textbf{Response to lighting interventions.} In each sequence one source (B) is brightened in steps relative to a second source (A) at a fixed query, crossing the point where B becomes dominant. Pre and Post are the accuracies of the decoded dominant source before and after the crossing, Never and Premature are the fractions of sequences in which the decoded source never switches to B after the crossing or switches before it, Stable is the fraction of pre-crossing frame pairs whose decoder margin keeps its sign, and Share MAE and Response are the error of the attribution-head shares against the rendered shares and the fraction of steps whose share change has the correct sign.}
\label{tab:intervention}
\resizebox{\linewidth}{!}{%
\begin{tabular}{@{}l ccc cc c cc@{}}
\toprule
& \multicolumn{3}{c}{Decoded dominant source (\%)} & \multicolumn{2}{c}{Switch failures (\%) \(\downarrow\)} & Stable & \multicolumn{2}{c}{Attribution head} \\
\cmidrule(lr){2-4}\cmidrule(lr){5-6}\cmidrule(lr){8-9}
Model & Pre & Post & All & Never & Premature & (\%) \(\uparrow\) & Share MAE \(\downarrow\) & Response (\%) \(\uparrow\) \\
\midrule
\rowcolor{oursgray} Ours & 56.7 & 46.2 & 51.4 & 40.9 & 47.7 & 61.9 & 0.220 & 87.3 \\
\bottomrule
\end{tabular}%
}
\end{table}

\subsubsection{Ablation Experiment Details}
\paragraph{Attribution and position head.}
The auxiliary objectives have only modest effects on dominant-source localization, but their differences become more visible at later source ranks. Contribution-share supervision improves rank-2 recovery over CE-only training, while the full objective gives the strongest rank-3 results and the highest rank-1 PCK at the tightest 0.25\,m threshold. Position regression alone does not improve these localization metrics, suggesting that its benefit is not well captured by PCK in isolation.
\begin{table}[h]
\centering
\small
\setlength{\tabcolsep}{4pt}
\renewcommand{\arraystretch}{1.05}
\caption{\textbf{Training-objective ablation by source rank (Infinigen test set).} 3D PCK (\%) of the predicted source at rank \(r\) for the models of Table~\ref{tab:ablation_objective}; \(N_q\) counts queries with at least \(r\) ground-truth sources.}
\label{tab:ablation_slots}
\begin{tabular}{@{}l r ccc ccc ccc@{}}
\toprule
& & \multicolumn{3}{c}{Rank 1 (\(N_q\)=1{,}032)} & \multicolumn{3}{c}{Rank 2 (\(N_q\)=586)} & \multicolumn{3}{c}{Rank 3 (\(N_q\)=169)} \\
\cmidrule(lr){3-5}\cmidrule(lr){6-8}\cmidrule(lr){9-11}
Training objective & & 0.25\,m & 0.5\,m & 1.0\,m & 0.25\,m & 0.5\,m & 1.0\,m & 0.25\,m & 0.5\,m & 1.0\,m \\
\midrule
CE only & & 47.3 & 69.7 & 78.2 & 29.7 & 45.9 & 51.0 & 31.4 & 37.3 & 42.0 \\
CE + attribution & & 48.3 & 71.3 & 82.8 & 29.0 & 49.5 & 57.2 & 24.9 & 37.9 & 45.0 \\
CE + position & & 45.0 & 68.5 & 79.3 & 24.2 & 41.8 & 51.4 & 23.1 & 32.5 & 37.9 \\
\rowcolor{oursgray} Full & & 49.2 & 70.6 & 82.2 & 31.4 & 49.5 & 56.8 & 32.0 & 40.2 & 46.2 \\
\bottomrule
\end{tabular}
\end{table}

\subsubsection{Why the Irradiance Field Is a Separate Network}
\label{app:vlm_radiometry}
Before adding the dense network we tested whether absolute radiometric quantities can be read out of the VLM itself. 
All tests used the frozen structured model of Section~\ref{sec:learning_framework}, the base Qwen3-VL-8B for comparison, and render-verified per-query log-irradiance targets on Infinigen scenes, with pass criteria fixed before each test.
As a reference, rendering the probe from the true source radiometry with our renderer reproduces the ground-truth probe closely (median highlight error \(8^\circ\)), so the protocol itself is not the limitation.

We first fitted linear readouts of log-irradiance on the hidden state of the final prompt token.
The fine-tuned model reached a Spearman correlation of 0.28 across scenes and the base model 0.50, against a required 0.70, and neither improved on predicting the per-scene mean.
Pooling the visual tokens around the query marker instead of using the prompt token raised the correlation only to 0.43 and left the highlight direction of the rendered probe at \(21^\circ\), above the \(18^\circ\) criterion.
A small convolutional adapter that reads the raw image together with the model's structured output overfitted the 54 available scenes, reaching a training error near zero while the validation correlation stayed at 0.46.
Increasing the number of training scenes from 15 to 84 improved the relative brightness structure within a scene but left the absolute scale unchanged, with a persistent bias of about half a log unit.

Table~\ref{tab:app_vlm_radiometry} summarizes these tests. 

\begin{table}[h]
\centering\scriptsize\setlength{\tabcolsep}{3pt}
\caption{Reading absolute radiometry from the VLM. \(\rho\) is the Spearman correlation between predicted and ground-truth log-irradiance across scenes, MAE is in log units, and the highlight error is the angular error of the rendered probe's highlight. Criteria were fixed before each test.} 
\label{tab:app_vlm_radiometry}
\begin{tabular}{@{}p{0.40\linewidth} p{0.24\linewidth} p{0.22\linewidth} l@{}}
\toprule
Readout & Criterion & Result & Verdict \\
\midrule
Reference, true source radiometry rendered with our renderer & highlight \(<10^\circ\) & \(8^\circ\) & pass \\
Linear readout of the final prompt token (fine-tuned / base) & \(\rho\ge0.70\) & \(\rho\) 0.28 / 0.50 & fail \\
Linear readout of the visual tokens around the marker & \(\rho\ge0.70\), highlight \(\le18^\circ\) & \(\rho\) 0.43, \(21^\circ\) & fail \\
Convolutional adapter on the image and the structured output (54 scenes) & val \(\rho\ge0.56\), MAE \(\le0.65\) & \(\rho\) 0.46, MAE 0.72 (train 0.06) & fail \\
Scene-count curve, 15 to 84 scenes & MAE \(\le0.50\), \(\rho\ge0.56\) & MAE 0.55, \(\rho\) 0.55, bias \(-0.47\) & fail \\
\bottomrule
\end{tabular}
\end{table}

Two observations follow.
Direction is readable from the VLM, since the fine-tuned features beat the base features on direction (\(32^\circ\) against \(45^\circ\)), but magnitude is not, and the base features were even better than the fine-tuned ones for log-irradiance, so source-referenced fine-tuning does not create a radiometric representation at the token level.
The absolute scale also did not improve with more scenes, whereas within-scene relative structure did, so the token representation encodes relative brightness but not exposure-level magnitude.
A dense per-pixel regressor trained directly on rendered irradiance (Section~\ref{sec:rgb_irradiance}) provides the magnitude and color at every pixel with far denser supervision, and the VLM is kept to what it does well, naming the sources and their geometry.

\subsubsection{What the Source-Referenced Output Contributes to the Rendering}
\label{app:composition_ablation}
Table~\ref{tab:composition_ablation} keeps the dense irradiance map, its calibration, and the fill fraction fixed and varies only how the predicted sources are turned into emitters, on the 49 diffuse probes of the 17-scene held-out Infinigen set (low-sample renders without the look-at protocol).
A single predicted direction places one emitter along the dominant direction, and the composition places one emitter per ranked entry with power proportional to its contribution share; the ground-truth reference uses all ground-truth sources with their true shares.
We measure the direction of the shading gradient on the sphere as the angle between the centroid of the bright hemisphere of the rendered sphere and that of the reference sphere.
The composition halves the error of the single direction and approaches the ground-truth-source reference.
The remaining gap is dominated by errors in source identity and geometry, and even with ground-truth sources the error is not zero because every source is represented by a point emitter and a global fill.
\begin{table}[h]
\centering
\small
\setlength{\tabcolsep}{6pt}
\renewcommand{\arraystretch}{1.05}
\caption{\textbf{Emitter construction with a fixed irradiance map (Infinigen held-out probes).} Median shading-direction error over the 49 diffuse probes of the 17-scene held-out set (fill fraction 0.65).}
\label{tab:composition_ablation}
\begin{tabular}{@{}l c@{}}
\toprule
Emitters & Shading dir.\ (\(^\circ\)) \(\downarrow\) \\
\midrule
Single predicted direction & 45.1 \\
\rowcolor{oursgray} Share-weighted composition (Ours) & 23.7 \\
\midrule
All ground-truth sources and shares & 17.7 \\
\bottomrule
\end{tabular}
\end{table}

\end{document}